\documentclass[letterpaper]{article} 
\usepackage{aaai2027}
\usepackage{diagbox}
\usepackage{multirow}

\usepackage[hyphens]{url}  
\usepackage{graphicx} 
\usepackage{natbib}  
\usepackage{caption} 
\usepackage{algorithm}
\usepackage{algorithmic}
\usepackage{amsmath}

\usepackage{tabularx}
\usepackage{array}

\newcolumntype{Y}{>{\centering\arraybackslash}X}
\usepackage{newfloat}
\usepackage{listings}
\DeclareCaptionStyle{ruled}{labelfont=normalfont,labelsep=colon,strut=off} 
\floatstyle{ruled}
\newfloat{listing}{tb}{lst}{}
\floatname{listing}{Listing}

\usepackage{booktabs}

 \nocopyright

\title{AttnLink: Turning Attention into Schema Links for Text-to-SQL}
\author{
    Jinwang Song\equalcontrib,
    Tao Liu\equalcontrib,
    Haowen Zheng\equalcontrib,
    Xiangheng Li,
    Yifan Li,
    Hongying Zan
}
\affiliations{

}

\begin{document}

\maketitle

\begin{abstract}
Schema linking is a critical component of Text-to-SQL systems, but existing approaches often trade off contextual modeling capacity, score-based controllability, and inference efficiency. We introduce \textbf{AttnLink}, an attention-based framework that converts LLMs’ internal attention into continuous relevance scores for schema items. AttnLink extracts the attention from the generation-start position to candidate schema spans, enabling all candidates to be ranked in a single prefill pass without autoregressive decoding. We develop two variants: \textbf{AttnLink-U}, which directly probes pretrained attention without parameter updates, and \textbf{AttnLink-S}, which aligns the attention distribution with gold schema items through direct supervision. To improve coverage of multiple relevant schema items, AttnLink-S combines a set-mass objective with an adaptive probability-floor regularizer. The resulting scores support post-hoc precision--recall control through temperature scaling and cumulative-mass selection. Experiments on Spider, BIRD, and Spider2-SQLite show that AttnLink-S achieves mAP scores of 99.22\%, 95.95\%, and 83.29\%, respectively, with millisecond-scale schema-linking latency. It also yields the best or tied-best execution accuracy for downstream SQL generation in seven of nine generator–dataset settings. Code will be made available at https://github.com/Songjw133/AttnLink.
\end{abstract}

\section{Introduction}

Text-to-SQL translates natural-language questions into executable SQL queries, enabling users to access relational databases without writing SQL~\cite{hong2025next,shi2025survey,liu2025survey}. Despite rapid progress with LLMs, reliable generation remains challenging because models must jointly interpret user intent, database structure, and schema-specific terminology. A central component is \emph{schema linking}, which identifies the tables and columns needed to answer a question. Accurate schema linking reduces irrelevant context and strengthens question--schema grounding, thereby mitigating column-selection and join-path errors~\cite{li2023resdsql,cao2024rslsql}.

An effective schema linker should satisfy three requirements: \textbf{semantic capacity} to model complex and compositional question--schema relationships, \textbf{continuous controllability} through ranked relevance scores that adapt to different schema budgets, and \textbf{serving efficiency} without auxiliary models or autoregressive decoding. Existing LLM-based approaches typically trade off these properties. Generative methods prompt or fine-tune LLMs to output relevant schema items as text~\cite{pourreza2023din,gao2024dail,talaei2024chess}. Although semantically expressive, they produce discrete sets with limited post-hoc control over recall, precision, and schema budget. Retrieval-based methods rank schema items using embeddings or cross-encoders~\cite{li2023resdsql}. They provide continuous scores, but separate embedding or cross-encoder models may offer weaker contextual and compositional modeling than LLMs.

We propose \textbf{AttnLink}, an attention-based schema-linking framework that addresses all three requirements by converting an LLM's internal attention into continuous relevance scores for schema items. Our key observation is that, before any schema item or SQL token is decoded, the generation-start position---which we term the \emph{generation anchor}—already exhibits a contextual attention pattern over the prompt, including candidate schema spans. AttnLink uses this signal to rank all candidates simultaneously, combining LLM semantic modeling with controllable continuous ranking. The resulting scores support post-hoc precision--recall adjustment through temperature scaling and top-$p$ selection. Requiring only a single standard prefill pass, AttnLink avoids both auxiliary embedding/reranking models and autoregressive decoding, enabling efficient schema pruning within existing LLM serving infrastructure.

We instantiate AttnLink in two complementary variants. \textbf{AttnLink-U} extracts schema-relevance scores from a selected attention layer and head without parameter updates, providing a lightweight, high-recall pruning solution. \textbf{AttnLink-S} directly supervises the candidate attention distribution with a set-mass objective and an adaptive floor regularizer. The set-mass objective shifts probability toward relevant schema items, while the floor regularizer prevents less salient but necessary items from being overlooked in multi-positive examples. Both variants share the same scoring and inference procedure.

Experiments on Spider, BIRD, and Spider2-SQLite demonstrate that AttnLink provides strong schema-ranking quality, supports flexible precision--recall trade-offs, and improves downstream SQL generation through schema filtering. Our contributions are summarized as follows:

\begin{figure*}[htb!]
  \centering
  \includegraphics[width=0.95\linewidth]{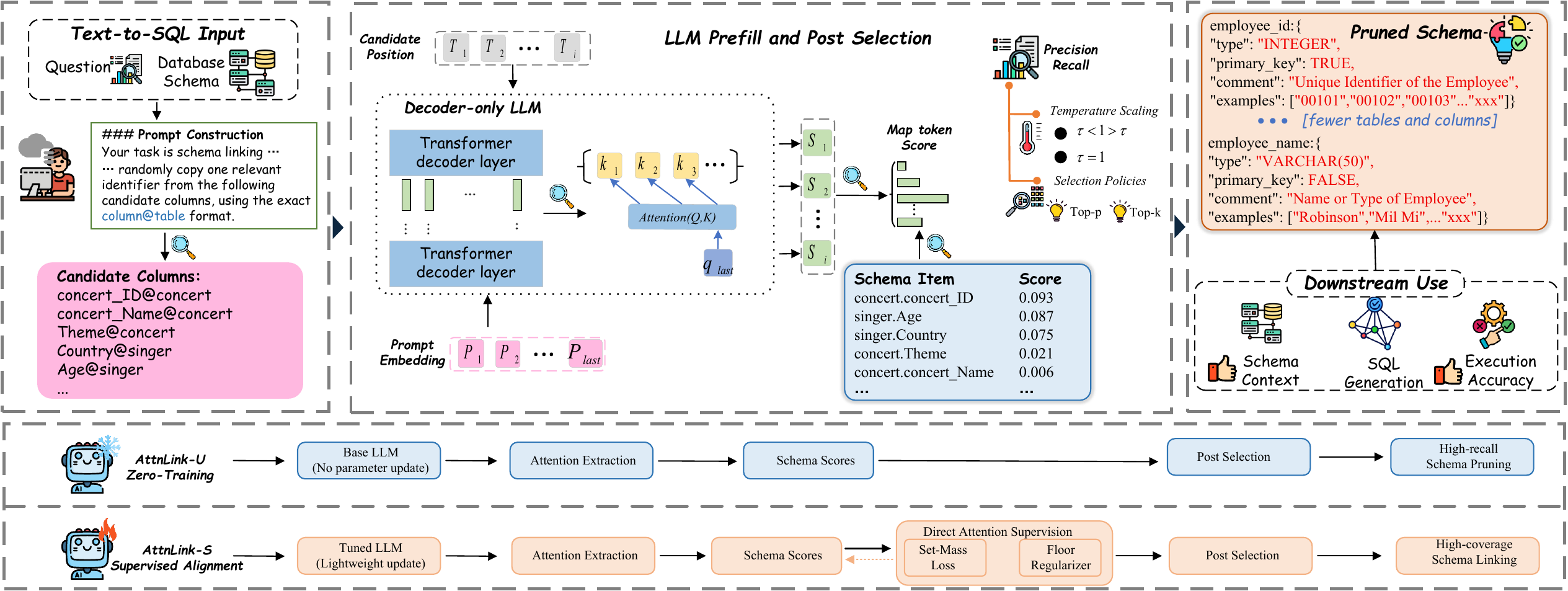}
  \caption{Overview of AttnLink.}
  \label{fig:main_pipeline}
\end{figure*}

\begin{enumerate}
\item We propose \textbf{AttnLink}, a unified schema linking framework that converts LLMs' internal attention into continuous table and column relevance scores. This formulation supports both zero-training probing and supervised alignment, and enables post-hoc precision--recall control through temperature scaling and top-$p$ cumulative-mass selection.
\item We show that LLM attention provides a strong and trainable schema grounding signal. AttnLink-U extracts useful schema links from base models without parameter updates, while AttnLink-S introduces a direct attention-alignment objective with set-mass loss and floor regularization to improve multi-positive coverage.
\item We validate AttnLink on BIRD, Spider, and Spider2-SQLite, demonstrating strong ranking quality, efficient single-prefill-pass inference without decoding, and compatibility across LLM architectures. AttnLink-S achieves the best or tied-best execution accuracy in seven of nine generator--dataset settings.

\end{enumerate}

\section{Problem Formulation}
\label{sec:problem_formulation}

Given a natural-language question $x$ and its associated database schema $\mathcal{S}$, schema linking aims to identify the tables and columns required to generate the correct SQL query. Depending on the linking granularity, we construct either a table-level or column-level candidate set from $\mathcal{S}$:
\begin{equation}
\mathcal{C} = \{c_1, c_2, \ldots, c_n\},
\end{equation}
where each $c_i$ denotes a candidate schema item. A schema-linking method models the semantic relationship between $x$ and $\mathcal{S}$ and predicts a relevant subset
\begin{equation}
\widehat{\mathcal{G}} \subseteq \mathcal{C}.
\end{equation}
Let $\mathcal{G} \subseteq \mathcal{C}$ denote the gold set of schema items required by the target SQL query. An ideal prediction should cover all items in $\mathcal{G}$ while excluding as many irrelevant candidates as possible.

Recall is particularly important in schema linking. Once a table or column required by the target SQL is removed, the downstream generator has little opportunity to recover it. By contrast, retaining a small number of irrelevant schema items usually introduces only limited contextual redundancy. Schema linking should therefore prioritize high recall while controlling the size and precision of the predicted subset.

\section{Method}
AttnLink converts the LLM’s internal attention at the generation anchor into continuous relevance scores over candidate schema items. As illustrated in Figure~\ref{fig:main_pipeline}, both variants share the same pipeline. AttnLink-U directly probes pretrained attention without parameter updates, while AttnLink-S further aligns the attention distribution with gold schema items through supervision. The resulting scores support flexible precision--recall control and efficient prefill-only schema linking.

\label{sec:method}

\subsection{AttnLink-U: Training-Free Attention Probing}
\label{sec:attnlink_u}

AttnLink-U treats the internal attention of a pretrained LLM as a
training-free probe for schema relevance. When the model is instructed
to copy a relevant schema item from the context, the underlying
retrieval process is reflected in the attention assigned to the
corresponding candidate span. AttnLink-U extracts this signal from a
selected attention head and converts it into a continuous relevance
score for each candidate schema item, without updating any model
parameters.

\paragraph{Copying-Prompt Construction.}
Given a question $x$, a database schema $\mathcal{S}$, and a candidate set $\mathcal{C}$, we construct a prompt containing, in order, a task instruction, the question, the complete schema, and a line-separated candidate list. The list represents each schema item in a canonical identifier format (e.g., \texttt{col@table}), enabling unambiguous span mapping and output parsing. The LLM is instructed to copy exactly one relevant identifier and output no SQL or other text. When multiple items are relevant, it is instructed to choose one at random rather than defaulting to the most likely or most confidently predicted item.

\begin{figure}[htb!]
  \centering
  \includegraphics[width=0.95\linewidth]{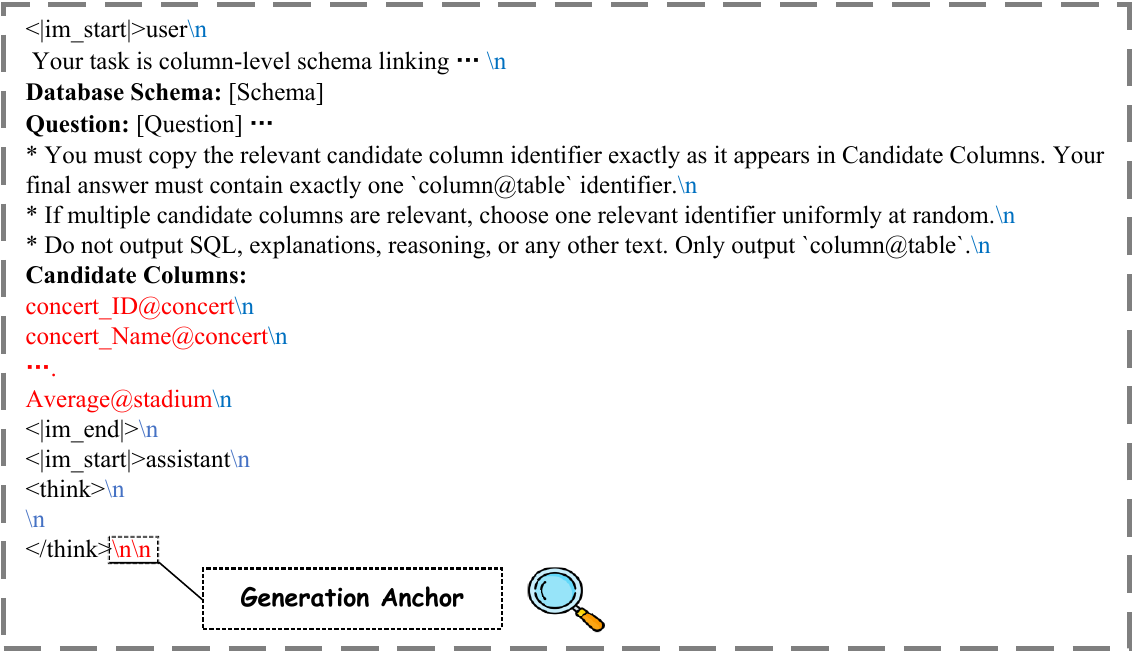}
  \caption{An illustration of the Generation Anchor.}
  \label{fig:teaser}
\end{figure}

We refer to the final token of the input prompt as the \emph{generation anchor}, as shown in Figure~\ref{fig:teaser}. Intuitively, because the LLM hidden state at this position is used to predict the first token of the copied identifier, the generation anchor should attend strongly to the span containing a relevant candidate and incorporate its identifying information into the current representation. Relevant schema items are therefore expected to receive larger anchor-to-candidate attention weights than irrelevant items.

We view next-token prediction at the generation anchor as a retrieval operation over candidate schema items. The copying prompt is used only to induce the desired attention pattern; AttnLink does not use the generated text as its prediction. Consequently, a single prefill pass is sufficient, and no candidate name needs to be autoregressively decoded.

\paragraph{Attention extraction at the generation anchor.}
Let $t$ denote the generation-anchor position. For layer $l$ and
attention head $h$, let $\mathbf{q}_{t}^{(l,h)}$ and
$\mathbf{k}_{j}^{(l,h)}$ denote the effective query and key vectors
used by the model, including any positional transformation or
normalization. In the canonical scaled dot-product form, the
self-attention weight from the generation anchor to a visible prefix
token $j$ is
\begin{equation}
a_{j}^{(l,h)}
=
\frac{
\exp\left(
\mathbf{q}_{t}^{(l,h)}
{\mathbf{k}_{j}^{(l,h)}}^{\top}
/ \sqrt{d_h}
\right)
}{
\sum_{r \leq t}
\exp\left(
\mathbf{q}_{t}^{(l,h)}
{\mathbf{k}_{r}^{(l,h)}}^{\top}
/ \sqrt{d_h}
\right)
},
\label{eq:anchor_attention}
\end{equation}
where $d_h$ is the head dimension. The denominator covers the entire
prompt prefix visible from the generation anchor, including the task
instruction, question, schema description, and candidate list.
Architecture-specific attention scaling is 
retained in implementation.

Different layers and attention heads encode different types of information, and not every head provides a reliable candidate-ranking signal. As shown in our experiments, strong ranking quality is concentrated in a subset of layers and heads. 

\paragraph{Candidate-Level Score Aggregation.}
A candidate schema item may consist of one or more tokens. Let $T_i$
denote the token span corresponding to candidate $c_i$ in the candidate
list. We aggregate the token-level attention over this span to obtain
the candidate score:
\begin{equation}
s_i
=
\operatorname{Pool}
\left(
\left\{
a_j \mid j \in T_i
\right\}
\right).
\label{eq:candidate_score}
\end{equation}
We use mean pooling by default to reduce the bias introduced by
differences in token-span length across schema items.

Because attention is also assigned to non-candidate tokens in the
prompt, the candidate scores do not by themselves define a normalized
distribution over $\mathcal{C}$. We therefore map them directly to a
temperature-scaled candidate distribution:
\begin{equation}
\pi_i^{(\tau)}
=
\frac{
\exp\left(
\log(s_i+\epsilon)/\tau
\right)
}{
\sum_{r=1}^{n}
\exp\left(
\log(s_r+\epsilon)/\tau
\right)
},
\label{eq:candidate_distribution}
\end{equation}
where $\epsilon$ is a small constant used for numerical stability and
$\tau>0$ controls the concentration of the distribution. When
$\tau=1$, Equation~\ref{eq:candidate_distribution} reduces to ordinary
candidate-wise normalization. The resulting distribution represents the relative relevance of each
candidate to the current question.

\subsection{AttnLink-S: Direct Attention Supervision}
\label{sec:attnlink_s}

AttnLink-S builds on AttnLink-U by retaining the same copying-style prompt, generation anchor, candidate-span pooling, and candidate-set normalization, while introducing direct supervision over the resulting candidate attention distribution. Given gold schema-linking annotations, it updates the LLM parameters to assign greater probability mass to the schema items required by the target SQL query, rather than relying solely on the pretrained attention pattern. During training, we use the unit-temperature distribution and write $\pi_i^{(1)}$ simply as $\pi_i$. Let
\begin{equation}
\mathcal{I}^{+}
=
\left\{
i \in \{1,\ldots,n\}
\mid
c_i \in \mathcal{G}
\right\}
\end{equation}
denote the indices of the gold schema items for a training example.
Because a question often requires multiple tables or columns, the
training objective must not only separate relevant from irrelevant
candidates, but also ensure that every positive candidate receives
sufficient probability mass.

\paragraph{Set-Mass Objective.}
We first define the total probability mass assigned to the gold positive set:
\begin{equation}
P_{\mathcal{G}}
=
\sum_{i \in \mathcal{I}^{+}} \pi_i.
\end{equation}
The corresponding set-mass loss is
\begin{equation}
\mathcal{L}_{\mathrm{set}}
=
-\log\left(P_{\mathcal{G}} + \epsilon\right).
\label{eq:set_mass_loss}
\end{equation}
This objective treats the positive candidates collectively, shifting probability mass from irrelevant candidates to the positive set without requiring a uniform allocation within the set.

A natural alternative is to optimize cross-entropy against a uniform target over the positives, assigning each a probability of $1/|\mathcal{I}^{+}|$. However, gold schema items may differ in their semantic association with the question, and enforcing uniformity can suppress these differences; empirically, this objective performs worse.

Nevertheless, using $\mathcal{L}_{\mathrm{set}}$ alone may cause
\emph{positive-set collapse}: a few easy positives can receive enough
probability mass to make the total positive mass large, while other
required items remain poorly scored, undermining the high recall
required for schema linking.

\paragraph{Adaptive probability floor.}
To improve coverage in multi-positive examples, we impose an adaptive lower bound on the probability of each positive candidate. For an example with $|\mathcal{I}^{+}|$ positives, we define the floor
\begin{equation}
f\!\left(\mathcal{I}^{+}; \rho\right)
=
\frac{\rho}{|\mathcal{I}^{+}|},
\label{eq:adaptive_floor}
\end{equation}
where $\rho \in [0,1]$ controls the total probability mass that the positive candidates are jointly encouraged to cover. The floor regularizer is
\begin{equation}
\mathcal{L}_{\mathrm{floor}}(\rho)
=
\frac{1}{|\mathcal{I}^{+}|}
\sum_{i \in \mathcal{I}^{+}}
\max
\left\{
0,\,
\log
\frac{
f\!\left(\mathcal{I}^{+}; \rho\right) + \epsilon
}{
\pi_i + \epsilon
}
\right\}.
\label{eq:floor_loss}
\end{equation}
A positive candidate is penalized only when its probability falls below
the adaptive floor; once the floor is reached, the corresponding
penalty becomes zero. The regularizer therefore prevents any required
schema item from receiving an excessively low score without forcing
all positive candidates to have equal probabilities.

The final training objective is
\begin{equation}
\mathcal{L}
=
\mathcal{L}_{\mathrm{set}}
+
\mathcal{L}_{\mathrm{floor}}(\rho).
\label{eq:attnlink_s_objective}
\end{equation}
Here, $\mathcal{L}_{\mathrm{set}}$ suppresses probability mass on irrelevant candidates, while $\mathcal{L}_{\mathrm{floor}}(\rho)$ prevents the distribution from concentrating on only a few positives. After training, AttnLink-S uses exactly the same inference procedure as AttnLink-U.

\subsection{Tunable Precision--Recall Trade-offs}
\label{sec:precision_recall_control}

Both AttnLink-U and AttnLink-S produce continuous candidate scores,
which Equation~\ref{eq:candidate_distribution} converts into a
normalized distribution $\boldsymbol{\pi}^{(\tau)}$. In contrast to
methods that directly generate a discrete list of schema items, this
distribution preserves both ranking and relative confidence
information. AttnLink can therefore adjust the predicted-set size
through temperature scaling and top-$p$ cumulative-mass selection,
without recomputing the model representations.

\paragraph{Temperature Scaling and Top-$\textbf{p}$ Selection.}
The temperature $\tau$ in
Equation~\ref{eq:candidate_distribution} preserves the candidate
ranking while redistributing probability mass across ranks. When
$\tau>1$, the distribution becomes flatter and assigns more probability
mass to lower-ranked candidates. When $\tau<1$, the distribution
becomes sharper and concentrates more mass on the highest-ranked
candidates.

We sort the candidates in descending order according to the
temperature-scaled distribution:
\begin{equation}
\pi_{(1)}^{(\tau)}
\geq
\pi_{(2)}^{(\tau)}
\geq
\cdots
\geq
\pi_{(n)}^{(\tau)}.
\end{equation}
For a cumulative-mass threshold $p\in(0,1]$, we select the shortest
ranked prefix whose cumulative probability reaches $p$:
\begin{equation}
k^{*}
=
\min
\left\{
k \in \{1,\ldots,n\}
:
\sum_{r=1}^{k}
\pi_{(r)}^{(\tau)}
\geq p
\right\}.
\label{eq:top_p_cutoff}
\end{equation}
The final prediction is
\begin{equation}
\widehat{\mathcal{G}}(p,\tau)
=
\left\{
c_{(1)},c_{(2)},\ldots,c_{(k^{*})}
\right\}.
\label{eq:top_p_prediction}
\end{equation}

\begin{table*}[t]
\centering
\begingroup
\small
\setlength{\tabcolsep}{1.4pt}
\renewcommand{\arraystretch}{1.10}
\setlength{\aboverulesep}{0.20ex}
\setlength{\belowrulesep}{0.20ex}
\begin{tabular*}{\textwidth}{
@{\extracolsep{\fill}}
l
@{\hspace{0.18cm}}
l|
*{4}{c}|
*{4}{c}|
*{4}{c}
@{}
}
\toprule
\textbf{Method}
&
\multicolumn{1}{l|}{\textbf{Model}}
&
\multicolumn{4}{c|}{\textbf{Spider Dev}}
&
\multicolumn{4}{c|}{\textbf{BIRD Dev}}
&
\multicolumn{4}{c}{\textbf{Spider2-SQLite}}
\\
\cmidrule(l{2pt}r{5pt}){3-6}
\cmidrule(l{5pt}r{5pt}){7-10}
\cmidrule(l{5pt}r{2pt}){11-14}
&
\multicolumn{1}{l|}{}
&
\textbf{SRR}
& \textbf{R}
& \textbf{P}
& \textbf{mAP}
&
\textbf{SRR}
& \textbf{R}
& \textbf{P}
& \textbf{mAP}
&
\textbf{SRR}
& \textbf{R}
& \textbf{P}
& \textbf{mAP}
\\
\midrule

%

\multicolumn{14}{c}{
\textbf{\textit{Embedding and Reranking Baselines}}
} \\
\cmidrule(lr){1-14}

Embedding
& BGE-M3
& 97.98 & 99.53 & 18.19 & 80.41
& 42.37 & 77.86 & 18.28 & 53.79
& 34.71 & 68.34 & 20.70 & 44.89
\\

Embedding
& Qwen3-Embedding-8B
& 95.87 & 98.86 & 18.04 & 78.83
& 38.07 & 72.34 & 16.96 & 49.13
& 34.71 & 70.14 & 20.87 & 48.03
\\

Reranking
& Qwen3-Reranker-4B
& 99.19 & 99.78 & 18.24 & 89.56
& 75.03 & 92.39 & 21.87 & 70.83
& 52.07 & 81.14 & 25.21 & 61.92
\\

Reranking
& Qwen3-Reranker-8B
& 99.29 & 99.82 & 18.25 & 89.68
& 72.23 & 92.43 & 21.92 & 71.75
& 52.89 & 81.05 & 25.50 & 63.11
\\

\midrule
\multicolumn{14}{c}{
\textbf{\textit{Generative SFT Linker}}
} \\
\cmidrule(lr){1-14}

DTS-SQL
& Qwen3.5-9B
& 92.36 & 97.76 & 95.12 & N/A
& 69.03 & 89.30 & \textbf{91.15} & N/A
& 35.56 & 66.73 & \textbf{77.93} & N/A
\\

\midrule
\multicolumn{14}{c}{
\textbf{\textit{Prompting/Agent-Based Linkers}}
} \\
\cmidrule(lr){1-14}

LinkAlign
& Qwen3.5-9B
& -- & -- & -- & N/A
& 68.51 & 87.87 & 87.55 & N/A
& 57.02 & 84.58 & 73.17 & N/A
\\

RSL-SQL
& Qwen3.5-9B
& -- & -- & -- & N/A
& 91.72 & 97.74 & 32.88 & N/A
& 78.51 & 93.87 & 34.56 & N/A
\\

AutoLink
& Qwen3.5-9B
& -- & -- & -- & N/A
& \textbf{98.44} & \textbf{99.60} & 11.69 & N/A
& \textbf{93.33} & \textbf{98.10} & 8.63 & N/A
\\

\midrule
\multicolumn{14}{c}{
\textbf{\textit{LLM-Based Precision--Recall Tunable Linkers}}
} \\
\cmidrule(lr){1-14}

ExSL
& Qwen2.5-Coder-7B
& 89.17 & 97.25 & 94.46 & 98.72
& 86.96 & 96.32 & 79.66 & 94.96
& -- & -- & -- & --
\\

JOLT-SQL
& Qwen2.5-Coder-7B
& 93.01 & 98.83 & \textbf{95.14} & 98.84
& 84.62 & 95.59 & 82.27 & 94.90
& 71.07 & 91.69 & 46.58 & 78.77
\\

\textbf{AttnLink-U}
& Qwen2.5-Coder-7B
& 99.90 & 99.96 & 18.56 & 88.36
& 90.15 & 97.64 & 19.51 & 75.82
& 66.11 & 88.20 & 21.10 & 59.09
\\

\textbf{AttnLink-U}
& Qwen3-4B
& 99.39 & 99.90 & 21.09 & 91.70
& 77.82 & 94.24 & 21.05 & 76.51
& 56.19 & 86.00 & 21.60 & 66.95
\\

\textbf{AttnLink-U}
& Qwen3.5-9B
& 99.79 & 99.95 & 21.12 & 92.27
& 91.38 & 98.01 & 23.02 & 79.84
& 61.98 & 87.11 & 24.80 & 67.80
\\

\textbf{AttnLink-U}
& Llama3.1-8B
& 99.80 & 99.96 & 17.75 & 87.05
& 85.45 & 96.49 & 19.38 & 73.15
& 76.86 & 94.70 & 15.45 & 55.11
\\

\textbf{AttnLink-U}
& Qwen3.5-35B-A3B
& \textbf{100} & \textbf{100} & 17.29 & 91.89
& 95.62 & 98.99 & 19.11 & 82.36
& 71.03 & 91.55 & 23.07 & 72.43
\\

\textbf{AttnLink-S}
& Qwen2.5-Coder-7B
& 94.58 & 99.49 & 86.39 & 98.96
& 96.15 & 99.14 & 45.84 & 95.25
& 76.21 & 94.03 & 38.85 & 81.25
\\

\textbf{AttnLink-S}
& Qwen3-4B
& 94.76 & 98.63 & 93.38 & 98.88
& 96.48 & 99.13 & 42.61 & 95.22
& 76.03 & 93.64 & 39.60 & 79.46
\\

\textbf{AttnLink-S}
& Qwen3.5-9B
& 97.02 & 99.25 & 94.27 & \textbf{99.22}
& 98.11 & \textbf{99.60} & 42.34 & \textbf{95.95}
& 91.11 & 97.10 & 34.05 & \textbf{83.29}
\\

\bottomrule
\end{tabular*}
\endgroup
\caption{
Column-level schema linking results on Spider Dev, BIRD Dev, and
Spider2-SQLite. For generative linkers that do not produce ranked
candidate lists, mAP is not applicable and is therefore reported as N/A.
}
\label{tab:schema-linking-results-fullwidth}
\end{table*}

Unlike nucleus sampling in language-model decoding, AttnLink retains the entire top-$p$ prefix rather than sampling a single item. Increasing $p$ or $\tau$ generally produces a larger set with higher recall, while decreasing them favors a more compact set with higher precision.

Top-$p$ selection adapts the output size to the concentration of AttnLink’s normalized relevance scores: it selects fewer candidates when the scores are concentrated and more when they are diffuse. A maximum-cardinality constraint (\texttt{top-p-max-k}) or a relative probability threshold (\texttt{min-p}) can be added to further control the candidate budget and remove low-scoring tail items.

\subsection{Efficient Prefill-Only Inference}
\label{sec:efficient_inference}

AttnLink computes all candidate relevance scores in a single prefill pass without autoregressive decoding. It preserves the model architecture, attention mask, and standard inference procedure, requiring only access to the Query--Key representations of a selected layer--head pair to compute generation-anchor attention. This design avoids the sequential decoding latency of generative schema linkers and integrates naturally with serving frameworks such as vLLM~\cite{kwon2023vllm}, benefiting from existing inference optimizations. In our experiments, the vLLM-based AttnLink implementation achieves millisecond-scale schema-linking latency.

\section{Experiments}
\subsection{Experimental Setup}
\label{sec}

\paragraph{Datasets.}

We evaluate AttnLink on three Text-to-SQL benchmarks: Spider~\cite{yu2018spider}, BIRD~\cite{li2023bird}, and the SQLite subset of Spider 2.0-Lite~\cite{spider2.0}.
\paragraph{Evaluation Metrics.}

For schema linking, we report Precision (\textbf{P}), Recall (\textbf{R}), Strict Recall Rate (\textbf{SRR}), and mean Average Precision (\textbf{mAP}). SRR measures complete gold-schema coverage, while mAP evaluates overall candidate-ranking quality across operating points. For downstream SQL generation, we report execution accuracy (\textbf{EX}).

\paragraph{Implementation Details.}

AttnLink uses a single layer--head pair for schema scoring, since strong ranking quality is concentrated in only a few heads and multi-head or cross-layer averaging yields no consistent gains. For AttnLink-U, we randomly sample 400 examples from the BIRD training set to select the best-performing layer--head pair for each model and reuse it across datasets. AttnLink-S instead supervises a head in the final layer and is trained on the available training split; since Spider2-SQLite does not provide a training split, the model trained on BIRD is directly transferred to Spider2-SQLite. We use top-$p=0.99$ and $\tau=2.0$ for both variants during inference, and set $\rho=0.25$ for AttnLink-S during training. Additional implementation details are provided in the supplementary materials.

\paragraph{Baselines.}
We compare AttnLink with several categories of schema-linking baselines: embedding-based retrievers, including BGE-M3~\cite{xiao2023bge} and Qwen3-Embedding-8B; cross-encoder rerankers, including Qwen3-Reranker-4B and Qwen3-Reranker-8B~\cite{qwen3-embedding}; the generative SFT method DTS-SQL~\cite{dts-sql}; prompting- or agent-based methods, including LinkAlign~\cite{linkalign}, RSL-SQL~\cite{cao2024rslsql}, and AutoLink~\cite{autolink}; and the LLM-based, precision--recall-tunable methods ExSL~\cite{glass2025exsl} and JOLT-SQL~\cite{song2025jolt}.

\subsection{Experimental Results}
\begin{figure*}[ht!]
  \centering
  \includegraphics[width=0.9\linewidth]{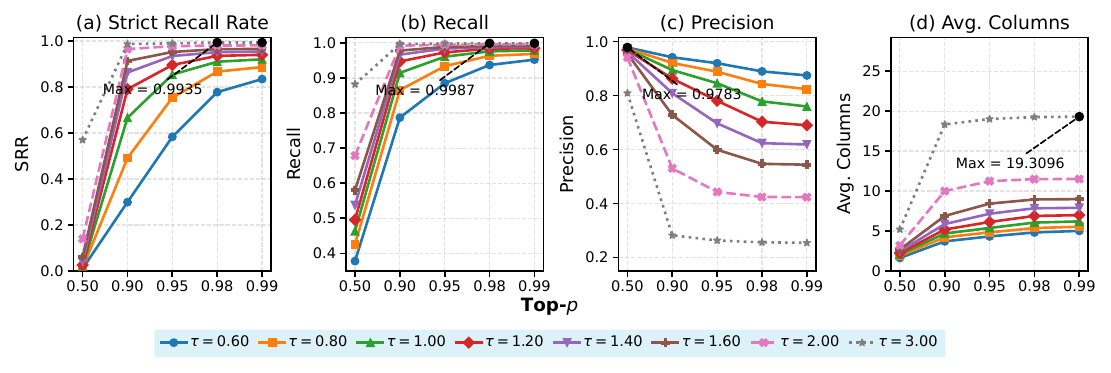}
  \caption{
    Temperature \& Top-$p$ sensitivity of Qwen3.5-9B AttnLink-S on BIRD Dev.
    }
  \label{fig:temperature_topp}
\end{figure*}

Table~\ref{tab:schema-linking-results-fullwidth} summarizes the column-level schema-linking results. AttnLink-S with Qwen3.5-9B achieves mAP scores of \textbf{99.22\%}, \textbf{95.95\%}, and \textbf{83.29\%} on Spider, BIRD, and Spider2-SQLite, respectively, setting state-of-the-art results among ranked linkers while retaining high SRR. Since Spider2-SQLite has no training split, the BIRD-trained model is directly transferred and still performs strongly, demonstrating robust cross-dataset generalization. Figure~\ref{fig:temperature_topp} shows that varying $p$ and $\tau$ smoothly controls the recall--precision--schema-size trade-off. The set-mass objective concentrates probability on gold items, while the floor regularizer preserves less salient but necessary ones. Without task-specific training, AttnLink-U remains effective across Qwen~\cite{qwen2_5_coder,qwen3,qwenteam2026qwen35omnitechnicalreport} and Llama~\cite{grattafiori2024llama} backbones spanning dense Transformers, hybrid-attention designs, and mixture-of-experts architectures, indicating that both the attention-based scoring mechanism and copying-prompt design generalize across model families and architectures.

\begin{table}[ht!]
\centering
\small
\setlength{\tabcolsep}{3.2pt}
\renewcommand{\arraystretch}{1.05}
\begin{tabular}{lcccccc}
\toprule
\multirow{2}{*}{\textbf{Dataset}}
& \multicolumn{6}{c}{$\rho$} \\
\cmidrule(lr){2-7}
& $0$ & $0.1$ & $0.25$ & $0.5$ & $0.75$ & $1.0$ \\
\midrule
BIRD Dev
& 88.41 & 95.54 & \textbf{95.95} & 95.72 & 95.48 & 94.96 \\
Spider2-SQLite
& 71.44 & 82.06 & \textbf{83.29} & 80.75 & 81.22 & 78.75 \\
\bottomrule
\end{tabular}
\caption{Ablation study of the AttnLink-S floor ratio $\rho$. Results are reported in mAP (\%).}
\label{tab:rho_ablation}
\end{table}

\subsection{Further Analysis}
\begin{figure}[ht!]
    \centering
    \includegraphics[width=\columnwidth]{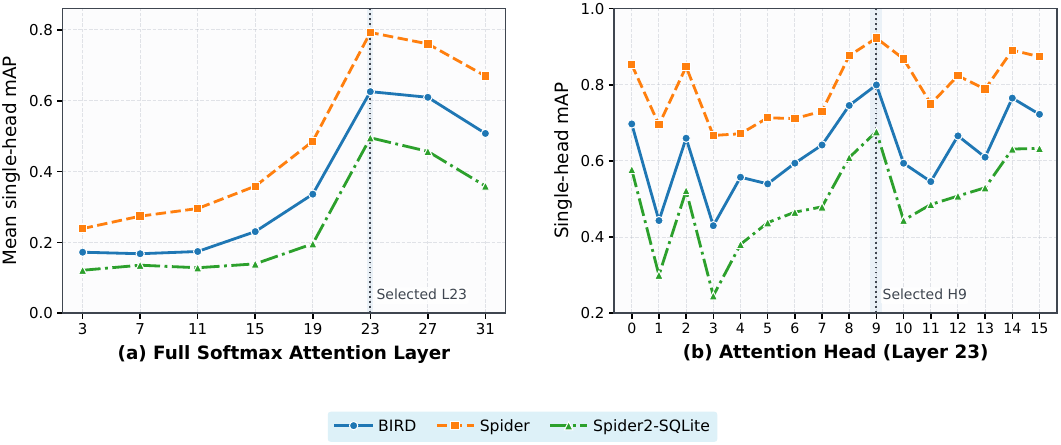}
    \caption{
    Layer--head analysis of AttnLink-U with Qwen3.5-9B. Panel (a) reports the mean single-head mAP for each full-attention layer, while panel (b) compares all attention heads in the selected layer (Layer 23).
}
    \label{fig:head_stability}
\end{figure}

\paragraph{Layer--Head Selection for AttnLink-U.}
\label{sec:layer_head_stability}
Figure~\ref{fig:head_stability} shows that retrieval quality varies sharply across layers and heads, making calibration necessary. We further observe that the strongest layer--head configurations tend to emerge in the middle-to-late layers, suggesting that schema-grounding signals become more pronounced after sufficient contextual and semantic processing. Despite differences in absolute mAP, evaluating the layer--head pairs separately on BIRD, Spider, and Spider2-SQLite identifies the same optimal pair, and the overall head rankings are strongly correlated across the three datasets. This cross-dataset stability suggests that certain attention heads specialize in retrieval-and-copy operations, capturing a transferable schema-grounding signal rather than dataset-specific patterns. Therefore, the layer--head configuration can be selected using a small held-out calibration set and subsequently reused across datasets.

\paragraph{Effect of the Floor Ratio for AttnLink-S.}
Table~\ref{tab:rho_ablation} shows that $\rho=0$, which uses only the set-mass objective, suffers from positive-set collapse and yields substantially lower mAP. A moderate floor ($\rho=0.25$) provides the best balance, whereas $\rho=1$ effectively recovers the target of uniform-positive cross-entropy: satisfying the floor $1/|I^+|$ for every positive candidate requires allocating all probability mass uniformly over the positive set. This prevents the model from emphasizing more salient schema items.

\paragraph{Efficiency Analysis.}

Table~\ref{tab:efficiency_main} reports concurrent inference latency for all methods using vLLM~0.21.0 on a single NVIDIA H100 80GB PCIe GPU. AttnLink uses the vLLM-Hook~\cite{ko2026vllmhook} framework to capture Q--K attention during prefill. For generative linkers, whose latency is dominated by autoregressive decoding, we additionally report time to last token (TTLT). Latency units are abbreviated as \textbf{ms}, \textbf{s}, \textbf{m}, and \textbf{h} for milliseconds, seconds, minutes, and hours, respectively. AttnLink preserves standard causal prefill computation. Unlike ExSL, which adds a classification head, and JOLT-SQL, which requires a bidirectional attention mask, AttnLink introduces no architectural or masking changes and remains compatible with existing serving frameworks. It therefore benefits from continuous batching, prefix caching, compilation, and PagedAttention, achieving millisecond-scale average latency.

\begin{table*}[hbt]
\centering
\begingroup

\small
\setlength{\tabcolsep}{1.0pt}
\renewcommand{\arraystretch}{1.00}
\setlength{\aboverulesep}{0.20ex}
\setlength{\belowrulesep}{0.20ex}

\begin{tabularx}{\textwidth}{
@{}
>{\raggedright\arraybackslash}p{1.60cm}
@{\hspace{0.10cm}}
>{\raggedright\arraybackslash}p{2.75cm}
*{12}{Y}
@{}
}
\toprule

\multirow{2}{*}{\textbf{Method}}
&
\multirow{2}{*}{\textbf{Model}}
&
\multicolumn{6}{c}{\textbf{BIRD Dev}}
&
\multicolumn{6}{c}{\textbf{Spider2-SQLite}}
\\

\cmidrule(lr){3-8}
\cmidrule(lr){9-14}

&
&
\shortstack{\textbf{Avg.}\\\textbf{Lat.}}
&
\shortstack{\textbf{Total}\\\textbf{Lat.}}
&
\shortstack{\textbf{Avg.}\\\textbf{TTLT}}
&
\shortstack{\textbf{Avg.}\\\textbf{Input}}
&
\shortstack{\textbf{Avg.}\\\textbf{Output}}
&
\shortstack{\textbf{Avg.}\\\textbf{Prefills}}
&
\shortstack{\textbf{Avg.}\\\textbf{Lat.}}
&
\shortstack{\textbf{Total}\\\textbf{Lat.}}
&
\shortstack{\textbf{Avg.}\\\textbf{TTLT}}
&
\shortstack{\textbf{Avg.}\\\textbf{Input}}
&
\shortstack{\textbf{Avg.}\\\textbf{Output}}
&
\shortstack{\textbf{Avg.}\\\textbf{Prefills}}
\\

\midrule

Embedding
& BGE-M3
& 38ms & 59s & -- & 1.0K & 0 & 47.1
& 81ms & 11s & -- & 1.9K & 0 & 68.1
\\

Embedding
& Qwen3-Embedding-8B
& 66ms & 1m42s & -- & 1.0K & 0 & 47.1
& 115ms & 15s & -- & 2.0K & 0 & 68.1
\\

Reranking
& Qwen3-Reranker-8B
& 157ms & 4m1s & -- & 8.7K & 0 & 75.6
& 237ms & 32s & -- & 17.6K & 0 & 105.7
\\

\cmidrule(l{4pt}r{4pt}){1-14}

DTS-SQL
& Qwen3.5-9B
& 90ms & 2m20s & 1s & 2.0K & 0.1K & --
& 162ms & 22s & 2s & 3.9K & 0.1K & --
\\

LinkAlign
& Qwen3.5-9B
& 15s & 6h22m & 4m28s & 12.7K & 18.9K & --
& 48s & 1h48m & 11m36s & 24.1K & 50.9K & --
\\

RSL-SQL
& Qwen3.5-9B
& 3s & 1h23m & 43s & 6.0K & 2.0K & --
& 6s & 13m59s & 54s & 7.6K & 2.7K & --
\\

AutoLink
& Qwen3.5-9B
& 4s & 1h43m & 1m33s & 32.9K & 2.5K & --
& 7s & 16m36s & 1m42s & 72.8K & 5.0K & --
\\

\cmidrule(l{4pt}r{4pt}){1-14}

\textbf{AttnLink}
& Qwen2.5-Coder-7B
& 15ms & 23s & -- & 2.0K & 0 & \textbf{1.0}
& 57ms & 8s & -- & 4.2K & 0 & \textbf{1.0}
\\

\textbf{AttnLink}
& Qwen3-4B
& \textbf{11ms} & \textbf{16s} & -- & 2.0K & 0 & \textbf{1.0}
& \textbf{37ms} & \textbf{5s} & -- & 4.2K & 0 & \textbf{1.0}
\\

\textbf{AttnLink}
& Qwen3.5-9B
& 32ms & 49s & -- & 2.1K & 0 & \textbf{1.0}
& 98ms & 13s & -- & 4.3K & 0 & \textbf{1.0}
\\

\textbf{AttnLink}
& Qwen3.5-35B-A3B
& 27ms & 41s & -- & 2.1K & 0 & \textbf{1.0}
& 67ms & 9s & -- & 4.3K & 0 & \textbf{1.0}
\\

\bottomrule
\end{tabularx}
\endgroup
\caption{
Schema linking inference efficiency on BIRD Dev and Spider2-SQLite.
Avg.\ Lat.\ denotes the average latency per example.
Input and output are average token counts per example, while prefills
denote the average prefill passes per example. 
}
\label{tab:efficiency_main}

\end{table*}

\paragraph{SQL Generation Performance.}

To isolate the effect of schema linking, we use a unified single-turn generation setting without self-consistency, voting, or iterative refinement. As shown in Table~\ref{table:sql_gen}, AttnLink-S achieves the best or tied-best EX in seven of nine generator--dataset settings, demonstrating that its ranking improvements translate into more reliable SQL generation rather than merely better intrinsic linking metrics. The gains are most pronounced on BIRD and Spider2-SQLite, where large and noisy schemas amplify two competing risks: missing required schema items and retaining excessive distractors. AttnLink-S addresses both by concentrating probability mass on the relevant set while preventing less salient gold items from being suppressed, thereby providing generators with schema contexts that are both complete and compact. AttnLink-U remains competitive without training, suggesting that pretrained attention already encodes transferable grounding signals, while direct supervision improves their calibration and discrimination. The optimal precision--recall balance is nevertheless generator-dependent: high SRR protects against irreversible omissions, whereas noise-sensitive generators benefit from sharper filtering.

\begin{table*}[t]
\centering
\begingroup
\small
\setlength{\tabcolsep}{1.7pt}
\renewcommand{\arraystretch}{1.10}

\begin{tabular*}{\textwidth}{
@{\extracolsep{\fill}}
l
l
*{10}{c}
@{}
}
\toprule

\multirow{2}{*}{\textbf{SQL Generator}}
&
\multirow{2}{*}{\textbf{Dataset}}
&
\multicolumn{10}{c}{\textbf{Schema Linking Method (EX, \char37)}}
\\

\cmidrule(lr){3-12}

&
&
\textbf{BGE-M3}
&
\shortstack{\textbf{Qwen3-}\\\textbf{Reranker-8B}}
&
\shortstack{\textbf{DTS-}\\\textbf{SQL}}
&
\shortstack{\textbf{RSL-}\\\textbf{SQL}}
&
\shortstack{\textbf{Auto}\\\textbf{Link}}
&
\shortstack{\textbf{Link}\\\textbf{Align}}
&
\textbf{ExSL}
&
\shortstack{\textbf{JOLT-}\\\textbf{SQL}}
&
\shortstack{\textbf{Attn}\\\textbf{Link-U}}
&
\shortstack{\textbf{Attn}\\\textbf{Link-S}}
\\

\midrule

\multirow{3}{*}{\textbf{Qwen3.5-9B}}
&
Spider Dev
& 83.4
& 83.8
& 83.9
& --
& --
& --
& \textbf{84.1}
& 83.8
& 84.0
& \textbf{84.1}
\\

&
BIRD Dev
& 49.3
& 64.6
& 67.7
& 68.0
& 67.8
& 66.1
& 67.1
& 67.5
& 67.2
& \textbf{69.1}
\\

&
Spider2-SQLite
& 7.4
& 9.1
& 10.7
& 14.9
& 14.9
& 15.7
& --
& 15.5
& 13.3
& \textbf{17.0}
\\

\midrule

\multirow{3}{*}{\textbf{Qwen3-4B}}
&
Spider Dev
& 85.0
& 85.4
& 84.8
& --
& --
& --
& 85.1
& 85.8
& 85.4
& \textbf{86.4}
\\

&
BIRD Dev
& 47.6
& 63.1
& 65.7
& 64.9
& 64.4
& 64.8
& 65.4
& 65.8
& 65.0
& \textbf{66.6}
\\

&
Spider2-SQLite
& 5.0
& 8.3
& 8.3
& \textbf{10.3}
& 6.6
& 8.3
& --
& 8.1
& 8.9
& 9.6
\\

\midrule

\multirow{3}{*}{
\textbf{\shortstack[l]{Qwen2.5-\\Coder-7B}}
}
&
Spider Dev
& 74.3
& 75.2
& 79.7
& --
& --
& --
& 79.6
& 80.2
& 77.2
& \textbf{80.6}
\\

&
BIRD Dev
& 42.1
& 52.1
& \textbf{59.1}
& 54.6
& 52.7
& 58.5
& 54.0
& 55.0
& 52.4
& 53.0
\\

&
Spider2-SQLite
& 4.4
& 3.0
& 6.6
& 6.6
& 5.9
& 5.0
& --
& 5.9
& 5.2
& \textbf{7.4}
\\

\bottomrule
\end{tabular*}
\endgroup
\caption{
Downstream SQL execution accuracy under different schema-linking methods. For both AttnLink-U and AttnLink-S, we use schema-linking outputs produced by the Qwen3.5-9B linker across all SQL generators.
}
\label{table:sql_gen}
\end{table*}


\section{Related Work}

\subsection{Schema Linking in Text-to-SQL}

Schema linking identifies the tables and columns required by a natural-language question and remains central to Text-to-SQL systems. Early neural approaches model question--schema relations, introduce schema-independent intermediate representations, or provide explicit linking supervision~\cite{wang2020ratsql,guo2019irnet,lei2020slsql}. Recent discriminative methods make schema selection explicit: RESDSQL~\cite{li2023resdsql} ranks schema items with a cross-encoder, while ExSL~\cite{glass2025exsl} and JOLT-SQL~\cite{song2025jolt} formulate schema linking as classification over LLM hidden states, requiring task-specific prediction heads and/or modifications to the causal attention mask. In parallel, generative and agent-based methods, including DIN-SQL~\cite{pourreza2023din}, DAIL-SQL~\cite{gao2024dail}, MAC-SQL~\cite{wang2024macsql}, CHESS~\cite{talaei2024chess}, DTS-SQL~\cite{dts-sql}, LinkAlign~\cite{linkalign}, RSL-SQL~\cite{cao2024rslsql}, and AutoLink~\cite{autolink}, select or refine schema items through prompting, generation, retrieval, or iterative exploration. These methods either produce discrete schema subsets with limited post-hoc precision--recall control or require prediction modules, retrieval stages, modified inference procedures, or autoregressive decoding.

\subsection{Attention as an Internal Signal}

Attention has long been studied as an internal model signal: individual heads encode structured relations, and task-relevant behavior often concentrates in a small subset of heads~\cite{clark2019bert,wiegreffe-pinter-2019-attention,bansal2023rethinking,yu2024large}. Recent work has further exploited attention-derived scores for efficient document and long-context retrieval~\cite{chen2025attention,zhang2025queryfocused}, yet attention-based schema linking remains largely unexplored. AttnLink targets this structured, multi-positive setting, where all relevant items must be jointly covered, assigned continuous relevance scores, and selected under variable schema budgets. It treats generation-anchor attention as an operational rather than causal signal, converting it into continuous candidate-level relevance scores in a single prefill pass and directly supervising the distribution for multi-positive coverage. To the best of our knowledge, AttnLink is the first Text-to-SQL schema linker to formulate schema linking directly over LLM attention, unifying training-free probing and supervised alignment within a shared inference framework while enabling post-hoc precision–recall control.
\section{Conclusion}
We present AttnLink, which converts LLM attention into continuous schema-relevance scores. AttnLink-U provides training-free attention probing, while AttnLink-S improves ranking and multi-positive coverage through set-mass supervision and an adaptive floor regularizer. AttnLink supports post-hoc precision--recall control and requires only a single prefill pass, achieving millisecond-scale inference without autoregressive decoding or an additional retrieval model. Experiments demonstrate strong schema-linking and downstream SQL performance, robust cross-dataset transfer, and consistent effectiveness across different LLM architectures.


\bibliography{main}


\newpage

\section{Appendix}
\section{Experimental Details}
\label{app}

\paragraph{Datasets and Gold-Schema Construction.}
We evaluate AttnLink on three Text-to-SQL benchmarks. For each example,
we reconstruct the candidate schema directly from the corresponding
SQLite database metadata and parse the reference SQL query using
\texttt{SQLGlot}. Table aliases are resolved, and all referenced
physical tables and columns are mapped back to their database-original
identifiers. The resulting sets are used as supervision for AttnLink-S
and as gold labels for schema-linking evaluation.

\begin{itemize}

\item \textbf{Spider}~\cite{yu2018spider} contains 7,000 training
examples and 1,034 development examples, with databases separated
across splits. It primarily evaluates cross-domain generalization to
unseen schemas and compositional SQL structures.

\item \textbf{BIRD}~\cite{li2023bird} contains 9,428 original training
examples and 1,534 development examples. Compared with Spider, it
features larger databases, richer database contents, and questions
requiring more realistic value grounding and domain knowledge. During
preprocessing, we find that the gold SQL queries of 477 training
examples fail to execute against their corresponding databases. We
remove these examples from the training split
for AttnLink-S. This filtering is applied only to training;
evaluation is conducted on the complete development set.

\item \textbf{Spider2-SQLite} is the SQLite subset of
Spider2.0-Lite~\cite{spider2.0} and contains 135 examples. It targets
substantially more complex enterprise-level database environments.
Since it provides no training split, the AttnLink-S model trained on
BIRD is directly transferred for evaluation.

\end{itemize}

\paragraph{Schema-Linking Evaluation Metrics.}
Let $M$ denote the number of evaluation examples. For the $m$-th
example, let $\mathcal{G}_m$ and $\widehat{\mathcal{G}}_m$ denote the
gold and predicted schema-item sets, respectively.

We compute precision as

\begin{equation}
\mathrm{P}
=
\frac{1}{M}
\sum_{m=1}^{M}
\frac{
\left|
\widehat{\mathcal{G}}_m
\cap
\mathcal{G}_m
\right|
}{
\left|
\widehat{\mathcal{G}}_m
\right|
}.
\label{eq:app-metric-precision}
\end{equation}

Recall is computed as

\begin{equation}
\mathrm{R}
=
\frac{1}{M}
\sum_{m=1}^{M}
\frac{
\left|
\widehat{\mathcal{G}}_m
\cap
\mathcal{G}_m
\right|
}{
\left|
\mathcal{G}_m
\right|
}.
\label{eq:app-metric-recall}
\end{equation}

Precision measures the proportion of selected schema items that are
relevant, whereas recall measures the proportion of gold schema items
retained by the linker.

Strict Recall Rate measures the proportion of examples for which the
prediction covers the complete gold schema:

\begin{equation}
\mathrm{SRR}
=
\frac{1}{M}
\sum_{m=1}^{M}
\mathbf{1}
\left[
\mathcal{G}_m
\subseteq
\widehat{\mathcal{G}}_m
\right].
\label{eq:app-metric-srr}
\end{equation}

Unlike recall, which gives partial credit when only a subset of the
required schema items is retrieved, SRR counts an example as correct
only when all gold tables or columns are retained. It therefore
directly captures whether schema pruning removes an item required for
downstream SQL generation.

For ranked schema linkers, we additionally report mean Average
Precision (mAP), a standard information-retrieval metric that rewards
relevant schema items appearing early in the ranking. Let
$c_{m,(1)}, \ldots, c_{m,(n_m)}$ denote the candidates for example $m$
in descending score order, and let
$y_{m,k}=\mathbf{1}[c_{m,(k)}\in\mathcal{G}_m]$. The Average Precision
of example $m$ is

\begin{equation}
\mathrm{AP}_m
=
\frac{1}{|\mathcal{G}_m|}
\sum_{k=1}^{n_m}
y_{m,k}
\left(
\frac{1}{k}
\sum_{r=1}^{k}
y_{m,r}
\right).
\end{equation}

Mean Average Precision is

\begin{equation}
\mathrm{mAP}
=
\frac{1}{M}
\sum_{m=1}^{M}
\mathrm{AP}_m.
\end{equation}

Although mAP serves a role similar to PR-AUC, AttnLink normalizes
scores independently within each example. Since candidate sets vary
in size and composition, scores are not directly comparable across
examples, making a global PR-AUC potentially misleading. mAP instead
evaluates each within-example ranking before averaging across the
dataset. It is not applicable to generative linkers that return only
an unranked subset.

\subsection{Implementation Details}
\label{app:implementation_details}

\paragraph{AttnLink-S Training.}
We train AttnLink-S using LoRA for parameter-efficient optimization.
The LoRA adapters are applied to all linear layers of the backbone
LLM, while the remaining model parameters are frozen. Table
\ref{tab:attnlink_training_hyperparameters} summarizes the training
configuration. We use an 8-bit AdamW optimizer and BF16 mixed-precision
training. A micro-batch size of one and four gradient-accumulation
steps give an effective batch size of four.

\begin{table}[t]
\centering
\small
\setlength{\tabcolsep}{4pt}
\renewcommand{\arraystretch}{1.08}
\begin{tabularx}{\columnwidth}{@{}lX@{}}
\toprule
\textbf{Hyperparameter} & \textbf{Value} \\
\midrule
LoRA target modules & All linear layers \\
LoRA rank $r$ & 64 \\
LoRA scaling $\alpha$ & 512 \\
LoRA dropout & 0.08 \\
Training epochs & 1 \\
Optimizer & 8-bit AdamW \\
Learning rate & $2 \times 10^{-5}$ \\
Learning-rate schedule &
Cosine annealing, $\eta_{\min}=5 \times 10^{-6}$ \\
Micro-batch size & 1 \\
Gradient accumulation & 4 steps \\
Effective batch size & 4 \\
Weight decay & $10^{-2}$ \\
Maximum gradient norm & 10.0 \\
Numerical precision & BF16 \\
\bottomrule
\end{tabularx}
\caption{Training hyperparameters for AttnLink-S.}
\label{tab:attnlink_training_hyperparameters}
\end{table}

As a representative example, training AttnLink-S with Qwen3.5-9B
takes approximately 45 minutes on a single NVIDIA H100 80~GB PCIe
GPU. Gradient checkpointing is not required on the H100. When
 gradient checkpointing is enabled, the same
configuration fits on a consumer GPU with 24~GB of memory and takes
approximately 90 minutes on a single NVIDIA RTX 4090.

\paragraph{Schema-Linking Inference.}
We perform schema-linking inference using vLLM~0.21.0 in BF16 precision,
with asynchronous request scheduling and TorchDynamo/Inductor
compilation enabled. We set
\texttt{max\_num\_seqs} to 256 and
\texttt{max\_num\_batched\_tokens} to 8192, and enable both prefix
caching and chunked prefill. The maximum model length is set to
100{,}000 tokens for the Qwen3/Qwen3.5 models and to 32{,}000 tokens for
Qwen2.5-Coder-7B, following the context-length limit of the Qwen2.5
backbone.

For hybrid-attention backbones, AttnLink probes only global
(full-attention) blocks, since linear-attention blocks do not expose
equivalent post-RoPE query and key representations. For example, the
global-attention blocks of Qwen3.5-9B are layers
$3, 7, 11, 15, 19, 23, 27,$ and $31$, using zero-based layer indices.

To retain compilation while extracting attention signals, we implement
a compilation-aware Q/K capture mechanism on top of
vLLM-Hook~\cite{ko2026vllmhook}. Specifically, the post-RoPE query and
key tensors are passed through an opaque, alias-preserving custom
operator, \texttt{vllm\_hook::capture\_qk}, immediately before the
vLLM attention operator. The capture operator is registered in
vLLM's \texttt{splitting\_ops}, which isolates its side-effecting
capture logic from the Inductor graph while preserving compiled
execution of the surrounding projection and MLP regions and retaining
vLLM's optimized attention kernels. Captured Q/K representations are
then processed asynchronously. In our warmed-run measurements, enabling Q/K capture introduces only
approximately 1.5\% additional latency relative to capture-disabled
vLLM, which is negligible compared with the millisecond-scale
inference latency of AttnLink.

\paragraph{SQL Generation.}
For downstream SQL generation, we likewise use vLLM and follow the
same basic vLLM serving configuration described above for
schema-linking inference. We use
stochastic decoding with \texttt{temperature}=0.7,
\texttt{top\_p}=0.8, \texttt{top\_k}=20, and
\texttt{max\_tokens}=8000. We fix the random seed throughout generation
to control sampling variability and ensure reproducibility.

\subsection{Baselines}
\label{app:baselines}

We compare AttnLink with the following schema-linking baselines:

\begin{itemize}

\item \textbf{Embedding Models.}
BGE-M3~\cite{xiao2023bge} and Qwen3-Embedding-8B~\cite{qwen3-embedding}
encode the question and each schema candidate independently. Candidates
are ranked according to the cosine similarity between their dense
representations, enabling efficient retrieval but without joint
question--candidate interaction.

\item \textbf{Qwen3 Rerankers.}
Qwen3-Reranker-4B and Qwen3-Reranker-8B~\cite{qwen3-embedding}
jointly process each question--candidate pair and produce a relevance
score. This pairwise interaction provides stronger contextual modeling
than independent embeddings, at the cost of a separate forward pass
for each candidate.

\item \textbf{DTS-SQL.}
DTS-SQL~\cite{dts-sql} decomposes Text-to-SQL into separately
fine-tuned schema-linking and SQL-generation stages. Its schema linker
autoregressively generates the relevant tables and columns as text,
producing a discrete schema subset rather than a ranked candidate list.

\item \textbf{LinkAlign.}
LinkAlign~\cite{linkalign} targets schema linking in large-scale and
multi-database settings through multi-round semantic retrieval,
irrelevant-information isolation, and schema-extraction enhancement.
These stages progressively narrow the schema before SQL generation.

\item \textbf{RSL-SQL.}
RSL-SQL~\cite{cao2024rslsql} combines bidirectional schema linking,
contextual information augmentation, binary schema selection, and
multi-turn self-correction. Its design prioritizes robust coverage of
required schema items while reducing irrelevant context.

\item \textbf{AutoLink.}
AutoLink~\cite{autolink} formulates schema linking as an iterative
agent-driven process. The agent autonomously explores and expands the
linked schema subset, allowing relevant items to be discovered without
placing the complete schema in every model call.

\item \textbf{ExSL.}
ExSL~\cite{glass2025exsl} adapts a decoder-only LLM to extractive
schema linking. It applies a task-specific classifier to candidate
hidden states to estimate relevance probabilities, providing
threshold-based control over the resulting precision--recall trade-off.

\item \textbf{JOLT-SQL.}
JOLT-SQL~\cite{song2025jolt} jointly optimizes discriminative schema
linking and SQL generation with a unified training objective. It uses
local bidirectional attention for schema classification and
confusion-aware noisy-schema sampling to improve robustness to
irrelevant schema items.

\end{itemize}

\section{Additional Analyses}

\subsection{Candidate-Order Stability of AttnLink-U}
\label{app:candidate-order-stability}

Since AttnLink-U derives candidate scores from positional attention
weights, we examine whether layer--head selection is sensitive to the
ordering of schema candidates. We randomly perturb the candidate order
eight times on BIRD Dev and recompute the
mAP of all 16 heads in Layer~23 of Qwen3.5-9B. The perturbations include
table-level, column-level, and joint table--column reordering.

\begin{figure*}[t]
\centering
\includegraphics[width=\textwidth]
{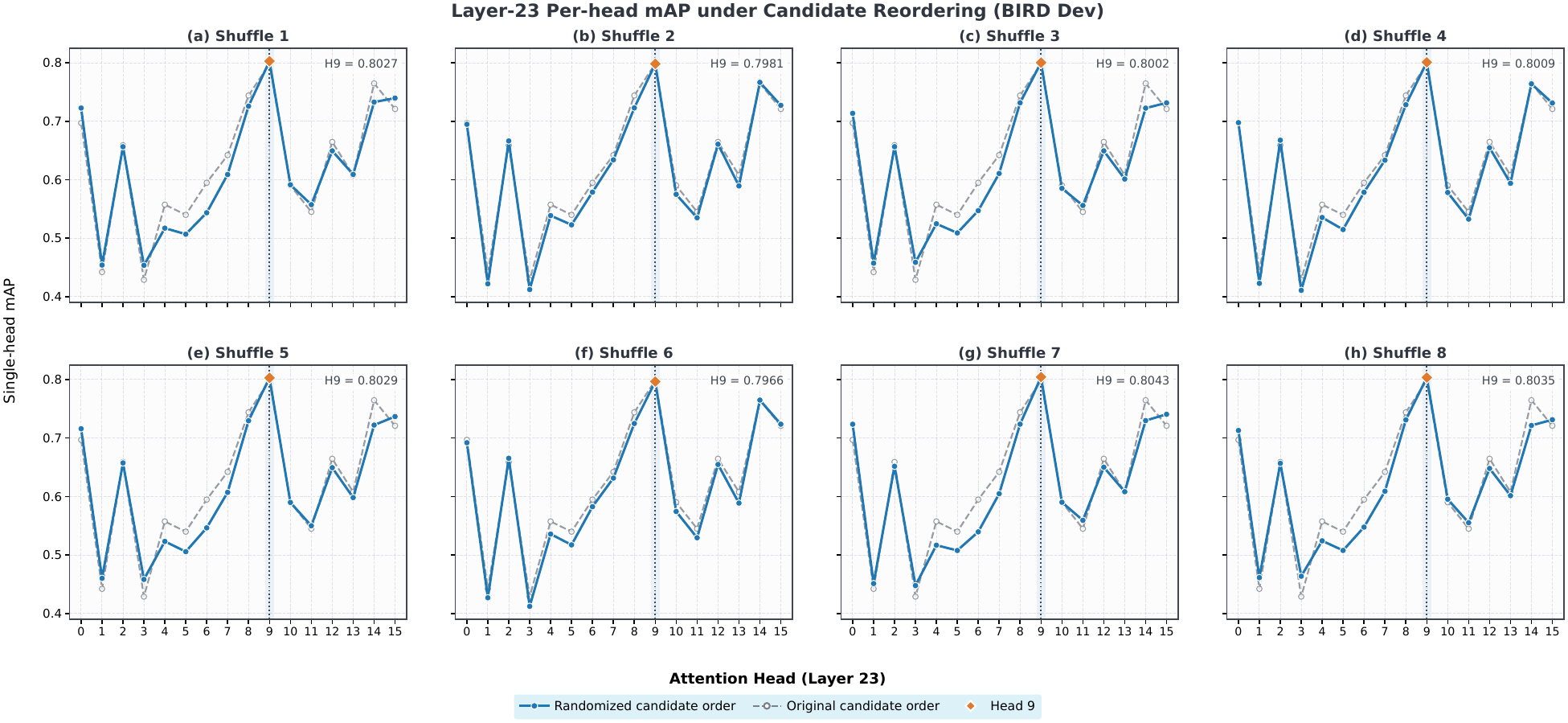}
\caption{
Per-head mAP of Qwen3.5-9B Layer~23 under eight randomized
candidate-order perturbations on BIRD Dev. The gray dashed curves
show results with the original candidate order, while the blue
curves show the corresponding shuffled results. Head~9, highlighted
by orange diamonds, remains the best-performing head in every run.
}
\label{fig:candidate-order-stability}
\end{figure*}

As shown in Figure~\ref{fig:candidate-order-stability}, Head~9 remains
the highest-performing head in all eight runs, with an average rank of
$1.00$. Its mAP is $0.8011 \pm 0.0027$, ranging only from $0.7966$ to
$0.8043$, compared with $0.7984$ under the original ordering. It also
maintains a clear advantage over the second-ranked head, with an
average margin of $0.0542$ and a minimum margin of $0.0316$.

The overall head ranking is similarly stable. The Spearman correlation
between each perturbed ranking and the original 16-head ranking is
$0.9794 \pm 0.0131$, with a minimum of $0.9676$. These results indicate
that candidate reordering may cause small fluctuations in absolute mAP,
but does not materially affect either the selected attention head or
the overall layer-level head ranking.

\subsection{Pooling Strategy}
\label{app:pooling_ablation}

AttnLink-U obtains each candidate-level score by aggregating
generation-anchor attention over the tokens of its identifier. Because
a candidate identifier may span multiple tokens, the pooling strategy
can directly affect ranking quality. We therefore examine whether
attention to the first identifier token is sufficient or whether
information should be aggregated over the complete candidate span.
Experiments use the fixed layer--head configuration selected for each
model: L23/H9 for Qwen3.5-9B, L22/H12 for Qwen2.5-Coder-7B, and
L24/H31 for Qwen3-4B. All conditions use the default
\texttt{col@table} candidate format. We compare attention to the first
identifier token (\texttt{first\_token}), mean pooling over the complete
span (\texttt{span\_mean}), and sum pooling over the complete span
(\texttt{span\_sum}).

\begin{table}[t]
\centering
\small
\setlength{\tabcolsep}{3.2pt}
\renewcommand{\arraystretch}{1.05}
\resizebox{\columnwidth}{!}{
\begin{tabular}{lcccc}
\toprule
\textbf{Model}
& \textbf{Layer/Head}
& \texttt{first\_token}
& \texttt{span\_mean}
& \texttt{span\_sum} \\
\midrule
Qwen3.5-9B
& L23/H9
& 78.09
& \textbf{79.84}
& 79.49 \\
Qwen2.5-Coder-7B
& L22/H12
& 71.91
& \textbf{75.82}
& 75.08 \\
Qwen3-4B
& L24/H31
& 74.72
& \textbf{76.51}
& 75.15 \\
\bottomrule
\end{tabular}
}
\caption{
Pooling ablation on BIRD Dev. All results use the
\texttt{col@table} candidate format and report mAP (\%).
}
\label{tab:pooling_ablation}
\end{table}

Despite the importance of the initial token,
Table~\ref{tab:pooling_ablation} shows that \texttt{span\_mean}
consistently achieves the highest mAP across all three models. Using
only \texttt{first\_token} reduces mAP by approximately $1.75$, $3.91$, and $1.79$ percentage points, respectively, indicating that later
tokens retain complementary identifier information. Although
\texttt{span\_sum} also uses the complete span, it remains consistently
below \texttt{span\_mean}, likely because summation introduces
sensitivity to differences in token-span length. These results support
the mean-pooling design used in our experiments, which captures the
complete identifier while controlling for candidate length.

\subsection{Necessity of the Copy-Oriented Instruction}
\label{app:copy_instruction_ablation}

Having established that AttnLink-U is stable to candidate reordering, we
next examine whether its ranking signal depends on the copy-oriented
instruction used in the main experiments. The default instruction asks
Qwen3.5-9B to copy exactly one relevant
identifier from the candidate list. We replace it with a
semantic-rationale instruction that asks the model to identify and
briefly explain all relevant columns using ordinary table and column
names, removing the exact-copy requirement, single-candidate output, and
random selection. The schema, question, candidate
set and order, and gold labels remain identical. Both settings use
Layer~23, Head~9, and \texttt{span\_mean} pooling without head
reselection.

\begin{table}[t]
\centering
\small
\setlength{\tabcolsep}{4.0pt}
\renewcommand{\arraystretch}{1.08}
\resizebox{\columnwidth}{!}{
\begin{tabular}{lcccc}
\toprule
\textbf{Dataset}
& \textbf{Copy-Oriented}
& \textbf{Semantic Rationale}
& \boldmath$\Delta$\unboldmath
& \textbf{Relative Change} \\
\midrule
BIRD
& \textbf{79.84}
& 25.61
& $-54.23$
& $-67.92\%$ \\
Spider
& \textbf{92.27}
& 34.72
& $-57.55$
& $-62.37\%$ \\
Spider2-SQLite
& \textbf{67.80}
& 21.65
& $-46.15$
& $-68.07\%$ \\
\bottomrule
\end{tabular}
}
\caption{
Copy-oriented instruction ablation for Qwen3.5-9B, reported in
mAP (\%). The copy-oriented values match the corresponding results in
the main comparison. $\Delta$ denotes semantic-rationale mAP minus
copy-oriented mAP in percentage points.
}
\label{tab:copy_instruction_ablation}
\end{table}

As shown in Table~\ref{tab:copy_instruction_ablation}, replacing the
copy-oriented instruction causes substantial and consistent
degradation. mAP decreases by $54.23$, $57.55$, and $46.15$ percentage
points on BIRD, Spider, and Spider2-SQLite, respectively, corresponding
to relative reductions of $62.37\%$--$68.07\%$. All paired sign-flip
tests yield $p<10^{-5}$, indicating that the degradation is
statistically significant. Thus, the copy-oriented prompting scheme is not merely an
output-formatting choice. By making a candidate identifier
the immediate generation target, it aligns the generation anchor with
the candidate spans and induces an attention distribution that is
useful for column-level ranking.

\begin{figure*}[h!]
\centering
\includegraphics[width=0.95\textwidth]
{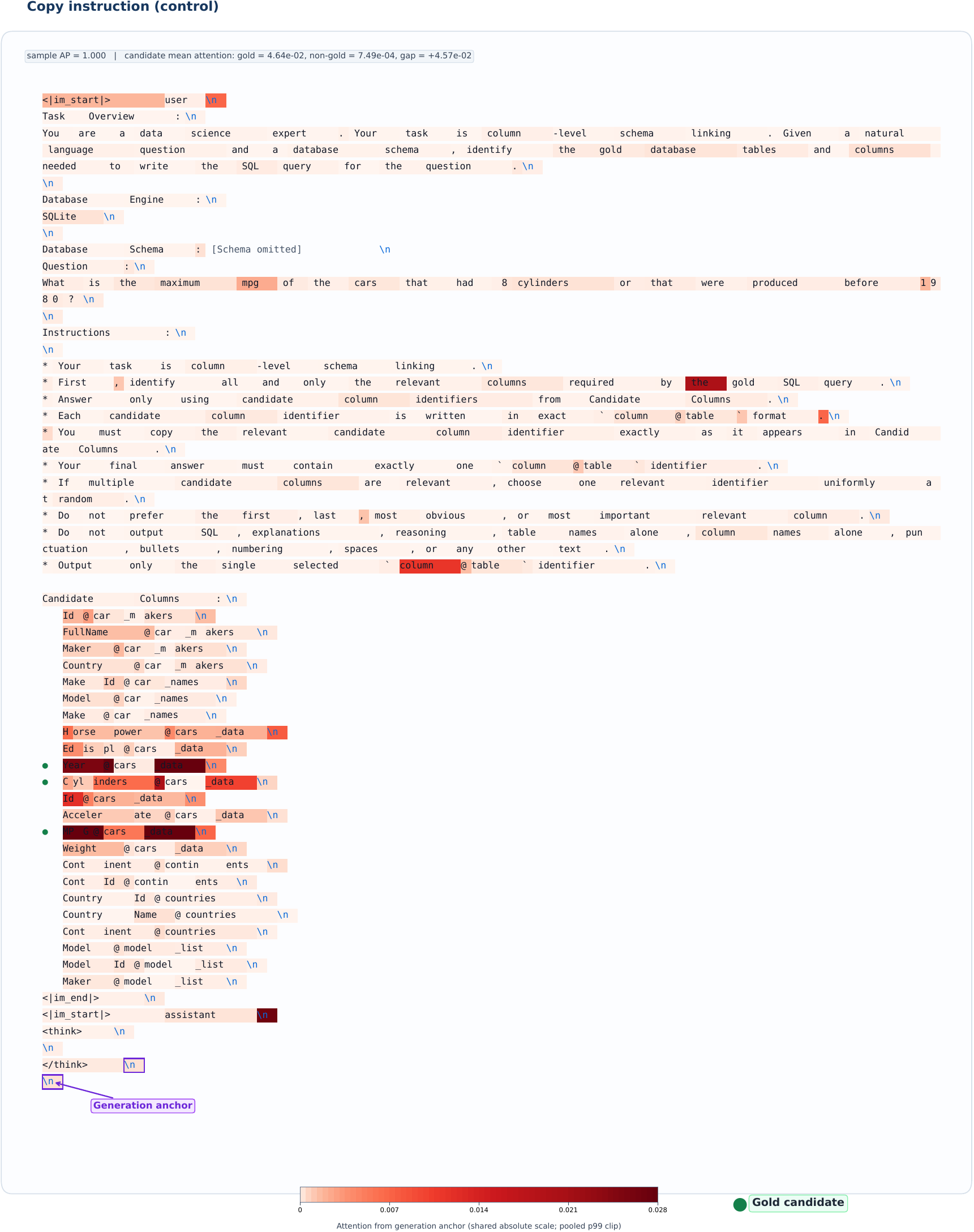}
\caption{
Generation-anchor attention of AttnLink-U under the copy-oriented
instruction for a Spider Dev example. Darker red indicates greater
attention from the generation anchor, and green dots mark gold
candidate columns.
}
\label{fig:copy_instruction_attention_control}
\end{figure*}

\begin{figure*}[h!]
\centering
\includegraphics[width=0.95\textwidth]
{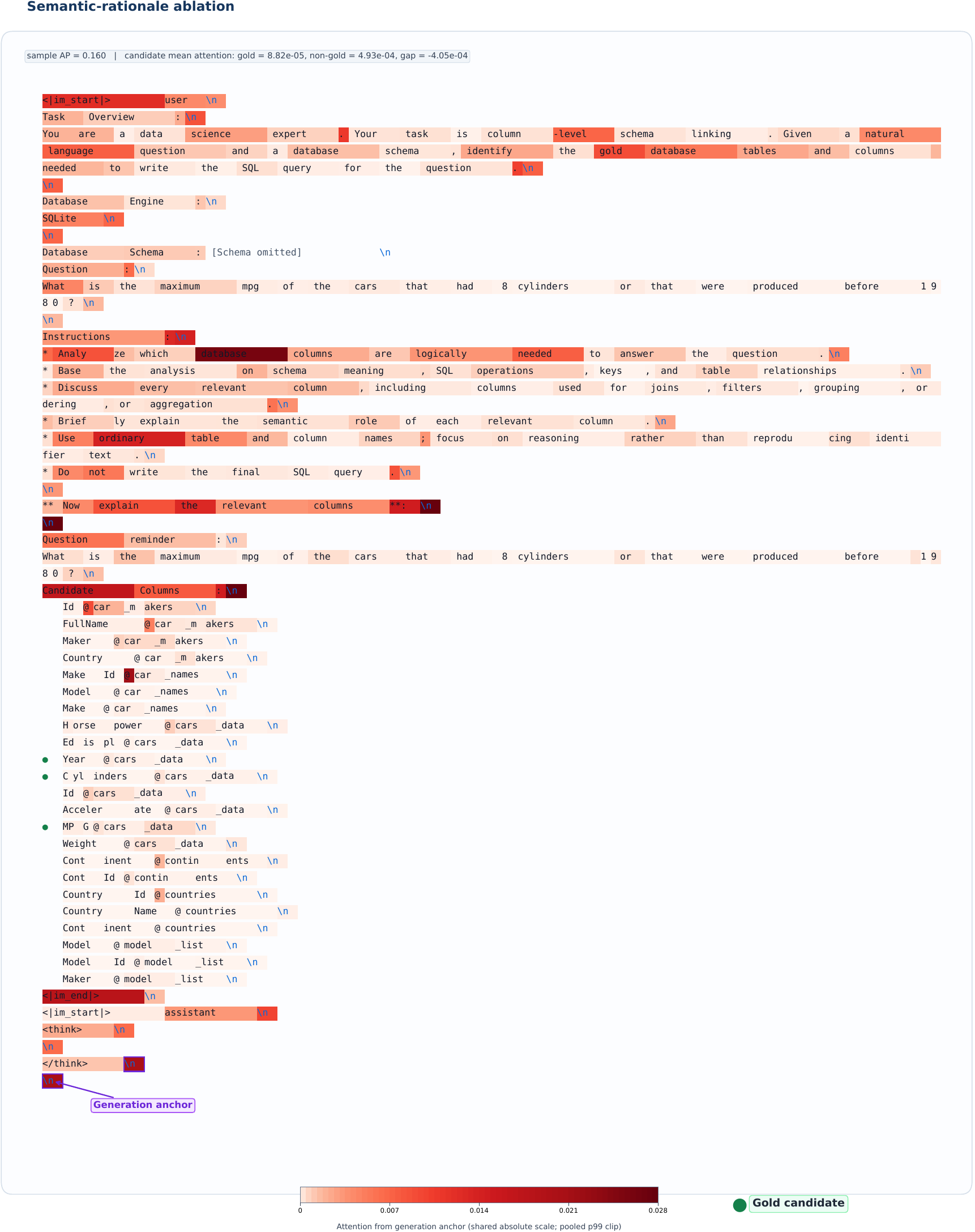}
\caption{
Generation-anchor attention of AttnLink-U for the same Spider example
after replacing the copy-oriented instruction with the
semantic-rationale instruction. The absolute color scale is shared
with Figure~\ref{fig:copy_instruction_attention_control}.
}
\label{fig:copy_instruction_attention_ablation}
\end{figure*}

Figures~\ref{fig:copy_instruction_attention_control}
and~\ref{fig:copy_instruction_attention_ablation} provide a qualitative
view of this effect. Under the copy-oriented instruction, the mean
attention assigned to gold candidates is $4.64\times10^{-2}$, compared
with $7.49\times10^{-4}$ for non-gold candidates. After ablation, the
corresponding means become $8.82\times10^{-5}$ and
$4.93\times10^{-4}$, respectively, eliminating and even reversing the
separation between gold and non-gold candidates. The example AP
accordingly decreases from $1.0000$ to $0.1597$. This comparison
supports the central design of AttnLink-U: the full copy-oriented prompting design is important for inducing a useful attention-based ranking signal, whereas generic semantic reasoning alone does
not reliably produce this behavior.

\subsection{Layer--Head Selection for AttnLink-S}
\label{app:attnlink-s-layer-head}

AttnLink-S directly supervises the attention distribution of a
selected layer--head pair. To examine its sensitivity to this choice,
we train Qwen3.5-9B AttnLink-S models using representative heads from
different layers while keeping all other training settings fixed.
Table~\ref{tab:attnlink-s-layer-head} reports the resulting mAP on
BIRD Dev.

\begin{table}[t]
\centering
\small
\setlength{\tabcolsep}{8.0pt}
\renewcommand{\arraystretch}{1.05}
\begin{tabular}{ccc}
\toprule
\textbf{Layer} & \textbf{Head} & \textbf{mAP (\%)} \\
\midrule
\textbf{31} & \textbf{0} & 95.95 \\
31 & 11 & 95.99 \\
27 & 0  & 96.02 \\
27 & 4  & 95.65 \\
23 & 15 & 96.01 \\
19 & 15 & 95.68 \\
19 & 8  & 95.44 \\
15 & 5  & 95.12 \\
15 & 8  & 94.84 \\
3  & 2  & 75.67 \\
3  & 9  & 76.60 \\
\bottomrule
\end{tabular}
\caption{
Layer--head sensitivity of Qwen3.5-9B AttnLink-S on BIRD Dev.
The configuration used in the main experiments is shown in bold.
}
\label{tab:attnlink-s-layer-head}
\end{table}

AttnLink-S is largely insensitive to the supervised head when the
target is placed in a middle-to-late layer. Across Layers~19--31, all
tested configurations achieve between $95.44\%$ and $96.02\%$ mAP,
while Layer~15 already reaches approximately $95\%$. These results
suggest that, after sufficient contextual processing, schema-grounding
signals can be reliably induced across different attention heads and
do not depend on a narrowly specialized layer--head pair.

In contrast, supervision at Layer~3 yields substantially lower mAP.
 Placing the supervision target too early therefore reduces
the effective trainable depth and limits the model's capacity to align
its attention with schema relevance. Given the small differences among
middle-to-late layers, we use Head~0 in the final layer throughout the
main experiments, avoiding additional layer--head selection.

\section{Additional Experimental Results}

\subsection{Tabular Temperature--Top-$p$ Sensitivity}
\label{app:temperature-top-p-table}

Table~\ref{tab:app-temperature-top-p} provides the complete
column-level results underlying the temperature--top-$p$ sensitivity
analysis in Figure~3 of the main paper.

\begin{table*}[t]
\centering
\begingroup
\small

\setlength{\tabcolsep}{4.0pt}
\renewcommand{\arraystretch}{1.06}

\newcommand{\metriccell}[4]{%
  \ensuremath{%
  #1\mkern4mu/\mkern4mu
  #2\mkern4mu/\mkern4mu
  #3\mkern4mu/\mkern4mu
  #4%
  }%
}

\newcommand{\metriccellbold}[4]{%
  \ensuremath{%
  \mathbf{#1}\mkern4mu/\mkern4mu
  \mathbf{#2}\mkern4mu/\mkern4mu
  \mathbf{#3}\mkern4mu/\mkern4mu
  \mathbf{#4}%
  }%
}

\begin{tabular*}{\textwidth}{
@{\extracolsep{\fill}}
c c c c
@{}
}

\toprule
\multicolumn{4}{c}{\textbf{Lower Top-$p$ Thresholds}}
\\
\midrule

\textbf{Temp. $\tau$}
&
\textbf{$p=0.50$}
&
\textbf{$p=0.75$}
&
\textbf{$p=0.90$}
\\
\midrule

$0.60$
& \metriccell{0.0104}{0.3780}{0.9783}{1.62}
& \metriccell{0.0756}{0.5897}{0.9662}{2.64}
& \metriccell{0.2992}{0.7869}{0.9421}{3.70}
\\

$0.70$
& \metriccell{0.0117}{0.4023}{0.9778}{1.74}
& \metriccell{0.0932}{0.6340}{0.9615}{2.87}
& \metriccell{0.3996}{0.8312}{0.9329}{3.96}
\\

$0.80$
& \metriccell{0.0124}{0.4261}{0.9745}{1.87}
& \metriccell{0.1108}{0.6688}{0.9571}{3.06}
& \metriccell{0.4909}{0.8654}{0.9209}{4.21}
\\

$0.90$
& \metriccell{0.0143}{0.4461}{0.9723}{1.97}
& \metriccell{0.1317}{0.6979}{0.9505}{3.23}
& \metriccell{0.5880}{0.8941}{0.9085}{4.45}
\\

$1.00$
& \metriccell{0.0189}{0.4635}{0.9719}{2.06}
& \metriccell{0.1604}{0.7251}{0.9446}{3.39}
& \metriccell{0.6662}{0.9145}{0.8945}{4.68}
\\

$1.10$
& \metriccell{0.0228}{0.4796}{0.9710}{2.14}
& \metriccell{0.1917}{0.7505}{0.9394}{3.55}
& \metriccell{0.7360}{0.9327}{0.8803}{4.91}
\\

$1.20$
& \metriccell{0.0254}{0.4958}{0.9687}{2.24}
& \metriccell{0.2373}{0.7766}{0.9320}{3.72}
& \metriccell{0.7907}{0.9470}{0.8627}{5.18}
\\

$1.30$
& \metriccell{0.0332}{0.5171}{0.9661}{2.34}
& \metriccell{0.2901}{0.8012}{0.9252}{3.89}
& \metriccell{0.8292}{0.9570}{0.8365}{5.49}
\\

$1.40$
& \metriccell{0.0391}{0.5374}{0.9617}{2.46}
& \metriccell{0.3507}{0.8244}{0.9164}{4.07}
& \metriccell{0.8631}{0.9664}{0.8088}{5.86}
\\

$1.50$
& \metriccell{0.0482}{0.5584}{0.9591}{2.57}
& \metriccell{0.4120}{0.8474}{0.9076}{4.27}
& \metriccell{0.8892}{0.9725}{0.7723}{6.32}
\\

$2.00$
& \metriccell{0.1395}{0.6785}{0.9410}{3.19}
& \metriccell{0.7425}{0.9398}{0.8112}{5.65}
& \metriccell{0.9648}{0.9919}{0.5302}{9.99}
\\

$3.00$
& \metriccell{0.5698}{0.8820}{0.8091}{5.20}
& \metriccell{0.9635}{0.9916}{0.4428}{13.14}
& \metriccell{0.9870}{0.9969}{0.2819}{18.33}
\\

\midrule
\multicolumn{4}{c}{\textbf{Higher Top-$p$ Thresholds}}
\\
\midrule

\textbf{Temp. $\tau$}
&
\textbf{$p=0.95$}
&
\textbf{$p=0.98$}
&
\textbf{$p=0.99$}
\\
\midrule

$0.60$
& \metriccell{0.5834}{0.8849}{0.9199}{4.31}
& \metriccell{0.7777}{0.9369}{0.8897}{4.81}
& \metriccell{0.8344}{0.9521}{0.8747}{5.01}
\\

$0.70$
& \metriccell{0.6825}{0.9131}{0.9036}{4.58}
& \metriccell{0.8351}{0.9532}{0.8682}{5.08}
& \metriccell{0.8664}{0.9615}{0.8504}{5.26}
\\

$0.80$
& \metriccell{0.7536}{0.9336}{0.8884}{4.82}
& \metriccell{0.8670}{0.9632}{0.8430}{5.36}
& \metriccell{0.8866}{0.9684}{0.8239}{5.53}
\\

$0.90$
& \metriccell{0.8103}{0.9487}{0.8696}{5.08}
& \metriccell{0.8866}{0.9692}{0.8133}{5.66}
& \metriccell{0.9022}{0.9732}{0.7940}{5.83}
\\

$1.00$
& \metriccell{0.8540}{0.9606}{0.8458}{5.38}
& \metriccell{0.9100}{0.9758}{0.7784}{6.05}
& \metriccell{0.9205}{0.9785}{0.7596}{6.20}
\\

$1.10$
& \metriccell{0.8774}{0.9684}{0.8162}{5.72}
& \metriccell{0.9224}{0.9796}{0.7424}{6.43}
& \metriccell{0.9270}{0.9804}{0.7261}{6.56}
\\

$1.20$
& \metriccell{0.8957}{0.9727}{0.7807}{6.13}
& \metriccell{0.9355}{0.9836}{0.7030}{6.88}
& \metriccell{0.9400}{0.9846}{0.6900}{6.98}
\\

$1.30$
& \metriccell{0.9198}{0.9792}{0.7421}{6.60}
& \metriccell{0.9420}{0.9852}{0.6623}{7.33}
& \metriccell{0.9446}{0.9861}{0.6553}{7.40}
\\

$1.40$
& \metriccell{0.9348}{0.9836}{0.6971}{7.16}
& \metriccell{0.9518}{0.9875}{0.6237}{7.84}
& \metriccell{0.9537}{0.9881}{0.6187}{7.90}
\\

$1.50$
& \metriccell{0.9452}{0.9862}{0.6487}{7.77}
& \metriccell{0.9589}{0.9893}{0.5843}{8.39}
& \metriccell{0.9602}{0.9895}{0.5808}{8.43}
\\

$2.00$
& \metriccell{0.9778}{0.9946}{0.4429}{11.23}
& \metriccell{0.9804}{0.9957}{0.4241}{11.49}
& \metriccell{0.9811}{0.9960}{0.4234}{11.51}
  \textsuperscript{$\dagger$}
\\

$3.00$
& \metriccell{0.9902}{0.9978}{0.2629}{19.01}
& \metriccell{0.9935}{0.9987}{0.2560}{19.25}
& \metriccell{0.9935}{0.9987}{0.2542}{19.31}
\\

\bottomrule
\end{tabular*}

\endgroup

\caption{
Temperature--top-$p$ sensitivity of Qwen3.5-9B AttnLink-S on BIRD
Dev. Each cell reports \textbf{SRR\,/\,R\,/\,P\,/\,Avg.\ Cols}.
SRR, recall, and precision are reported on a $[0,1]$ scale, while
Avg.\ Cols.\ denotes the average number of selected columns. The
configuration used in the main experiments is 
marked with $\dagger$.
}
\label{tab:app-temperature-top-p}
\end{table*}

Increasing either $\tau$ or $p$ improves schema coverage by retaining
more candidates, resulting in higher SRR and recall but lower
precision. The main configuration, $\tau=2.0$ and $p=0.99$, achieves
an SRR of $0.9811$ and a recall of $0.9960$ while selecting $11.51$
columns on average.

\subsection{SQL Generation Prompt}
\label{app:sql-generation-protocol}

\begin{figure*}[htbp]
  \centering
  \includegraphics[width=0.99\textwidth]
  {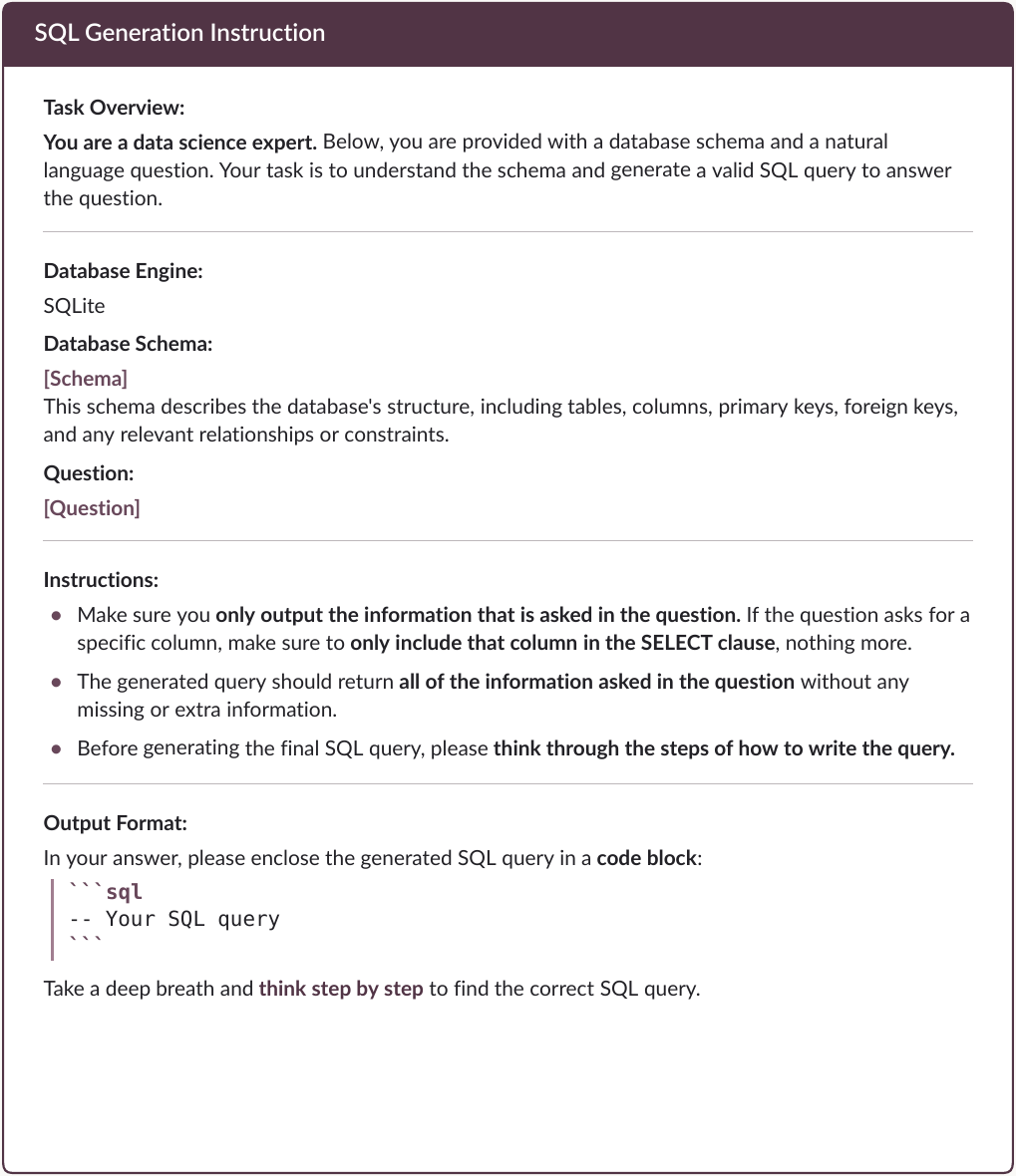}
  \caption{
  Single-turn prompt used for downstream SQL generation,
  following the prompt format of OmniSQL.
  }
  \label{fig:sql-generation-prompt}
\end{figure*}

Figure~\ref{fig:sql-generation-prompt} presents the unified prompt used
for downstream SQL generation. Following the prompt format of
OmniSQL~\cite{li2025omnisql}, it provides the database engine, the schema
retained by the evaluated linker, and the input question, together with
explicit generation instructions and a structured SQL output format.
For every example, we invoke the corresponding generator exactly once
and decode a single SQL query. We use the same prompt throughout all SQL-generation experiments, without self-consistency, majority voting, multi-turn
correction, or iterative refinement, ensuring that the reported EX
reflects the contribution of schema linking under a controlled
generation setting.

\subsection{Temperature Sensitivity of Downstream Execution Accuracy}
\label{app:ex-temperature-sensitivity}

\begin{figure*}[t]
  \centering
  \includegraphics[width=\textwidth]
  {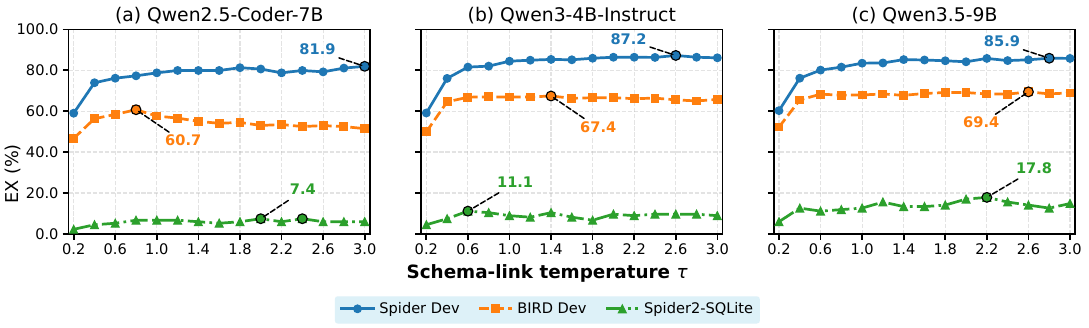}
  \caption{
  Downstream execution accuracy as a function of the schema-link
  temperature $\tau$. The Qwen3.5-9B AttnLink-S linker and
  $p=0.99$ are fixed, while the three panels correspond to different
  SQL generators. Annotated markers indicate the highest observed EX
  for each dataset and generator.
  }
  \label{fig:overall_ex_temperature_sensitivity}
\end{figure*}

Figure~\ref{fig:overall_ex_temperature_sensitivity} examines how
downstream execution accuracy changes with the schema-link temperature
$\tau$, while fixing the AttnLink-S linker, the top-$p$ threshold at
$p=0.99$, and all SQL-generation settings. Increasing $\tau$ flattens
the normalized relevance distribution, causing top-$p$ selection to
retain more candidates and thereby trade precision for recall. Across
most generator--dataset combinations, EX initially improves as
previously omitted schema items are recovered, but eventually plateaus
or declines as additional distractors enter the generator context.
The resulting curves therefore show that neither maximal precision nor
maximal recall is uniformly optimal; the preferred operating point
depends on both the SQL generator and the dataset.

The best observed EX values for Qwen2.5-Coder-7B are $\textbf{81.9\%}$,
$\textbf{60.7\%}$, and $\textbf{7.4\%}$ on Spider, BIRD, and Spider2-SQLite,
respectively; the corresponding values are $\textbf{87.2\%}$, $\textbf{67.4\%}$, and
$\textbf{11.1\%}$ for Qwen3-4B, and $\textbf{85.9\%}$, $\textbf{69.4\%}$, and $\textbf{17.8\%}$ for
Qwen3.5-9B. The clearest generator-dependent pattern appears on BIRD.
Qwen2.5-Coder-7B peaks at $\tau=0.8$, whereas Qwen3-4B peaks at $\tau=1.4$. Qwen3.5-9B instead reaches
its maximum at $\tau=2.6$, favoring a higher recall. This progression suggests that a stronger
generator can make effective use of a more complete but noisier schema
context, whereas a less robust generator may benefit more from
suppressing distractors. We view this as an empirical tendency rather
than a universal rule, since the Spider curves favor high recall for
all three generators and the smaller Spider2-SQLite benchmark exhibits
greater point-to-point variation.

This sensitivity analysis highlights a practical advantage of
AttnLink's continuous relevance scores: its precision--recall operating
point can be calibrated to a particular downstream generator without
retraining the linker. The benefit is especially pronounced for
Qwen2.5-Coder-7B on BIRD, where reducing $\tau$ from the main-paper
setting of $2.0$ to $0.8$ raises EX from $53.0\%$ to $\textbf{60.7\%}$, an
absolute gain of $7.7$ percentage points. Using the best observed
temperature for each generator--dataset pair improves or matches the
fixed setting in all nine cases. These calibrated results also exceed
the strongest non-AttnLink baseline reported in the main paper for
every generator--dataset combination, by approximately $0.8$--$2.1$
percentage points. Relative to all main-paper entries, including the
fixed AttnLink-S configuration, they establish a new maximum in eight
settings and tie the remaining Qwen2.5-Coder-7B Spider2-SQLite setting.
Although these post-hoc optima are reported as a sensitivity analysis
rather than as the uniform setting used for the main comparison, they
demonstrate that AttnLink can adapt schema coverage to the distinct
noise tolerance of different generators and convert that
controllability into measurable execution gains.

\subsection{Table-Level Schema-Linking Results}
\label{app:table-level-results}

We further evaluate AttnLink at the table level on Spider and BIRD, as shown in Table~\ref{tab:app-table-level-results}.

\begin{table*}[t]
\centering
\begingroup
\small
\setlength{\tabcolsep}{3.0pt}
\renewcommand{\arraystretch}{1.08}
\setlength{\aboverulesep}{0.20ex}
\setlength{\belowrulesep}{0.20ex}

\begin{tabular*}{\textwidth}{
@{\extracolsep{\fill}}
l|
*{4}{c}|
*{4}{c}
@{}
}
\toprule
\textbf{Model}
&
\multicolumn{4}{c|}{\textbf{Spider Dev}}
&
\multicolumn{4}{c}{\textbf{BIRD Dev}}
\\
\cmidrule(l{2pt}r{5pt}){2-5}
\cmidrule(l{5pt}r{2pt}){6-9}
&
\textbf{SRR}
& \textbf{R}
& \textbf{P}
& \textbf{mAP}
&
\textbf{SRR}
& \textbf{R}
& \textbf{P}
& \textbf{mAP}
\\
\midrule

\multicolumn{9}{c}{
\textbf{\textit{AttnLink-U}}
}
\\
\cmidrule(lr){1-9}

Qwen2.5-Coder-7B
& \textbf{99.61}
& \textbf{99.88}
& 54.83
& \textbf{98.00}
& \textbf{97.65}
& \textbf{99.14}
& 50.51
& 94.80
\\

Qwen3-4B
& 95.94
& 98.40
& \textbf{66.67}
& 96.71
& 91.00
& 96.15
& \textbf{76.73}
& 95.28
\\

Qwen3.5-9B
& 97.58
& 99.02
& 60.08
& 96.23
& 96.54
& 98.53
& 54.42
& 93.80
\\

Qwen3.5-35B-A3B
& 99.23
& 99.71
& 60.82
& 97.46
& 96.09
& 98.57
& 58.53
& \textbf{96.15}
\\

\midrule

\multicolumn{9}{c}{
\textbf{\textit{AttnLink-S}}
}
\\
\cmidrule(lr){1-9}

Qwen2.5-Coder-7B
& \textbf{99.90}
& \textbf{99.98}
& 89.10
& 99.91
& \textbf{98.11}
& \textbf{99.27}
& 72.62
& 98.26
\\

Qwen3-4B
& 99.71
& 99.92
& \textbf{93.47}
& \textbf{99.95}
& 97.46
& 99.07
& 76.01
& 98.36
\\

Qwen3.5-9B
& 99.71
& 99.92
& 92.21
& 99.92
& 98.04
& 99.18
& \textbf{81.70}
& \textbf{98.58}
\\

\bottomrule
\end{tabular*}
\endgroup

\caption{
Table-level schema-linking results on Spider Dev and BIRD Dev.
All metrics are reported as percentages (\%).
}
\label{tab:app-table-level-results}
\end{table*}

The table-level results are strong across model scales. Without
parameter updates, AttnLink-U reaches mAP scores of  $\textbf{98.00\%}$ on
Spider and  $\textbf{96.15\%}$ on BIRD, recalls of  $99.88\%$ and  $99.14\%$,
and SRRs of  $99.61\%$ and  $97.65\%$, respectively. This confirms that
pretrained attention already provides an effective table-linking
signal.

Direct attention supervision further pushes ranking quality close to
saturation. All three AttnLink-S backbones achieve at least  $99.91\%$
mAP,  $99.92\%$ recall, and  $\textbf{99.71\%}$ SRR on Spider. On BIRD, mAP
ranges from  $98.26\%$ to  $98.58\%$, with recall above  $99\%$ for every
backbone. Qwen3.5-9B achieves  $\textbf{98.58\%}$ mAP,  $81.70\%$ precision,
$99.18\%$ recall, and  $\textbf{98.04\%}$ SRR, while Qwen3-4B reaches the
highest Spider mAP of  $\textbf{99.95\%}$ with  $\textbf{93.47\%}$ precision, showing that
strong table-level performance does not depend on model scale alone.

\end{document}